\documentclass{article}

\usepackage{microtype}
\usepackage{graphicx}
\usepackage{subfigure}
\usepackage{booktabs} 

\usepackage{hyperref}

\usepackage[accepted]{icml2023}

\usepackage{amsmath}
\usepackage{amssymb}
\usepackage{mathtools}
\usepackage{amsthm}
\usepackage[capitalize,noabbrev]{cleveref}
\usepackage[font=small,skip=0pt]{caption}

\newcommand*\rot{\rotatebox{90}}
\usepackage{url}            
\usepackage{booktabs}       
\usepackage{amsfonts}       
\usepackage{nicefrac}       
\usepackage{microtype}      
\usepackage{xcolor}         

\usepackage{enumitem}
\usepackage{amsmath}

\usepackage{amsmath,amsfonts,bm}

\def\1{\bm{1}}

\DeclareMathAlphabet{\mathsfit}{\encodingdefault}{\sfdefault}{m}{sl}
\SetMathAlphabet{\mathsfit}{bold}{\encodingdefault}{\sfdefault}{bx}{n}

\newcommand{\E}{\mathbb{E}}

\DeclareMathOperator*{\argmin}{arg\,min}

\newcommand{\w}{\pmb{w}}
\theoremstyle{plain}
\newtheorem{theorem}{Theorem}[section]

\newtheorem{lemma}[theorem]{Lemma}

\theoremstyle{definition}
\newtheorem{definition}[theorem]{Definition}
\newtheorem{assumption}[theorem]{Assumption}
\theoremstyle{remark}

\usepackage[textsize=tiny]{todonotes}

\icmltitlerunning{Loss Balancing for Fair Supervised Learning}

\begin{document}
\setlength{\abovedisplayskip}{2pt}
\setlength{\belowdisplayskip}{2pt}
\setlength{\belowcaptionskip}{1pt}
\twocolumn[
\icmltitle{Loss Balancing for Fair Supervised Learning}




\begin{icmlauthorlist}
\icmlauthor{Mohammad Mahdi Khalili}{a,b}
\icmlauthor{Xueru Zhang}{b}
\icmlauthor{Mahed Abroshan}{c}
\end{icmlauthorlist}

\icmlaffiliation{a}{Yahoo Research, NYC, NY.}
\icmlaffiliation{b}{The Ohio State University, Columbus, OH.}
\icmlaffiliation{c}{Optum Labs, London, UK. The work was done while at Alan Turing Institute, London, UK}

\icmlcorrespondingauthor{Mohammad Mahdi Khalili}{khaliligarekani.1@osu.edu}

\icmlkeywords{Machine Learning, ICML}

\vskip 0.3in
]

\printAffiliationsAndNotice{} 
\begin{abstract}
 Supervised learning models have been  used in various domains such as lending, college admission, face recognition, natural language processing, etc. However, they 
 may inherit pre-existing biases from training data and exhibit discrimination against protected social groups. Various fairness notions have been proposed to address unfairness issues. In this work, we focus on Equalized Loss (EL), a fairness notion that requires the expected loss to be (approximately) equalized across different groups. Imposing EL on the learning process leads to a non-convex optimization problem even if the loss function is convex, and the existing fair learning  algorithms cannot properly be adopted   to find the fair predictor under the EL constraint. 
 This paper introduces an algorithm that can leverage off-the-shelf convex programming tools (e.g., CVXPY \cite{diamond2016cvxpy,agrawal2018rewriting}) to efficiently find the \textit{global} optimum of this non-convex optimization. In particular, we propose the $\texttt{ELminimizer}$ algorithm, which finds the optimal fair predictor under EL by reducing the non-convex optimization to a sequence of convex optimization problems. We theoretically prove that our algorithm finds the global optimal solution under certain conditions. Then, we support our theoretical results through several empirical studies.  
\end{abstract}

\section{Introduction}\label{sec:intro}
As machine learning (ML) algorithms are increasingly being used in applications such as education, lending, recruitment, healthcare, criminal justice, etc., there is a growing concern that the algorithms may exhibit discrimination against protected population groups. For example, speech recognition products such as Google Home and Amazon Alexa were shown to have accent bias \citep{accent}. The COMPAS recidivism prediction tool, used by courts in the US in parole decisions, has been shown to have a substantially higher false positive rate for African Americans compared to the general population \citep{COMPAS}.  Amazon had been using automated software since 2014 to assess applicants’ resumes, which were found to be biased against women \citep{amazon}. As a result, there have been several works focusing on interpreting machine learning models to understand how features and sensitive attributes contribute to the output of the model \cite{ribeiro2016lime,lundberg2017shap,abroshan2023symbolic}.  

Various fairness notions have been proposed in the literature to measure and remedy the biases in ML systems; they can be roughly classified into two categories: 1) \textit{individual fairness} focuses on equity at the individual level and it requires similar individuals to be treated similarly \citep{dwork2012fairness,biega2018equity,jung2019eliciting,gupta2019individual}; 2) \textit{group fairness} requires certain statistical measures to be (approximately) equalized across different groups distinguished by some sensitive attributes \citep{hardt2016equality,conitzer2019group,khalili2020improving,zhang2020long,khalili2021fair,diana2021minimax,williamson2019fairness,zhang2022fairness}. 

Several approaches have been developed to satisfy a given definition of fairness; they  fall under three categories: 1) \textit{pre-processing}, by modifying the original dataset such as removing certain features and reweighing, (e.g., \cite{kamiran2012data,celis2020data,abroshan2022counterfactual}); 2) \textit{in-processing},  by modifying the algorithms such as imposing fairness constraints or changing objective functions during the training process, (e.g., \cite{zhang2018mitigating,agarwal2018reductions,agarwal2019fair,reimers2021towards,calmon2017optimized}); 3) \textit{post-processing}, by adjusting the output of the algorithms based on sensitive attributes, (e.g., \cite{hardt2016equality}).   

In this paper, we focus on group fairness, and we aim to mitigate unfairness issues in supervised learning using an in-processing approach. This problem can be cast as a constrained optimization problem by minimizing  a loss function subject to a group fairness constraint. 
We are particularly interested in the Equalized Loss (EL)  fairness notion proposed by Zhang et al. \yrcite{zhang2019group},  which requires the expected loss (e.g., Mean Squared Error (MSE), Binary Cross Entropy (BCE) Loss) to be equalized across different groups.\footnote{ Zhang et al. \yrcite{zhang2019group} propose the EL fairness notion without providing an efficient algorithm for satisfying this notion.}  

The problem of finding fair predictors by solving constrained optimizations has been largely studied. Komiyama et al. \yrcite{komiyama2018nonconvex} propose the coefficient of determination constraint  for learning a fair regressor and develop an algorithm for minimizing the mean squared error (MSE) under  their proposed fairness notion.  Agarwal et al. \yrcite{agarwal2019fair} propose an approach to finding a fair regression model under bounded group loss and statistical parity fairness constraints. Agarwal et al. \yrcite{agarwal2018reductions} study classification problems and aim to find fair classifiers under various fairness notions including statistical parity and equal opportunity. In particular, they consider zero-one loss as the objective function and train a \textit{randomized} fair classifier over a finite hypothesis space. They show that this problem can be reduced to a problem of finding the saddle point of a linear Lagrangian function. 
 Zhang et al. \yrcite{zhang2018mitigating} propose an adversarial debiasing technique to find fair classifiers under equalized odd, equal opportunity, and statistical parity. Unlike the previous works, we focus on the \textit{Equalized Loss} fairness notion which has not been well studied. Finding an EL fair predictor  requires solving a non-convex optimization. Unfortunately, there is \underline{no algorithm} in fair ML literature with \underline{a theoretical performance guarantee} that can be properly applied to EL fairness (see Section \ref{sec:formulation} for detailed discussion).
 
 Our main contribution can be summarized as follows,
 \begin{itemize}[leftmargin=*,topsep=-1ex,itemsep=-0.2ex]
 \item We develop an algorithm with a theoretical performance guarantee for EL fairness. In particular, we propose the $(\texttt{ELminimizer})$ algorithm to solve a non-convex constrained optimization problem that finds the optimal  fair predictor under {EL} constraint. We show that such a non-convex optimization problem can be reduced to a sequence of  \textit{convex constrained} optimizations. The proposed algorithm finds the \textit{global optimal} solution and is applicable to both regression and classification problems. Importantly, it can be easily implemented using off-the-shelf convex programming tools. 
  \item In addition to $\texttt{ELminimizer}$ which finds the global optimal solution, we develop a simple algorithm for finding a \textit{sub-optimal}  predictor satisfying EL fairness. We prove there is a sub-optimal solution satisfying EL fairness that is a linear combination of the optimal solutions to two \textit{unconstrained} optimizations, and it can be found without solving any constrained optimizations.
 \item We conduct sample complexity analysis and provide a generalization performance guarantee. In particular, we show the sample complexity analysis found in \cite{donini2018empirical} is applicable to learning problems under EL. 
\item We also examine (in the appendix) the relation between Equalized Loss ({EL}) and Bounded Group Loss ({BGL}), another fairness notion proposed by \cite{agarwal2019fair}. We show that under certain conditions, these two notions are closely related, and they do not contradict each other.
  \end{itemize}

\section{Problem Formulation}
\label{sec:formulation}
Consider a supervised learning problem where the training dataset consists of triples $(\pmb{X},A,Y)$ from two social groups.\footnote{We use bold letters to represent vectors. } Random variable $\pmb{X} \in \mathcal{X}$ is the feature vector (in the form of a column vector), $A\in \{0,1\}$ is the sensitive attribute (e.g., race, gender) indicating the group membership, and $Y\in \mathcal{Y} \subseteq\mathbb{R}$ is the label/output.
We denote realizations of random variables by small letters (e.g.,  $(\pmb{x},a,y)$ is a realization of $(\pmb{X},A,Y)$). Feature vector $\pmb{X}$ may or may not include sensitive attribute $A$. Set $\mathcal{Y}$ can be either $\{0,1\}$ or $\mathbb{R}$: if $\mathcal{Y} = \{0,1\}$ (resp. $\mathcal{Y} = \mathbb{R}$), then the problem of interest is a binary classification (resp. regression) problem. 

Let $\mathcal{F}$ be a set of predictors $f_{\pmb{w}} : \mathcal{X}\to \mathbb{R}$ parameterized by weight vector $\pmb{w}$ with dimension $d_{\w}$.\footnote{Predictive models such as logistic regression, linear regression, deep learning models, etc., are parameterized by a weight vector.} 
If  the problem is binary classification, then $f_{\w}(\pmb{x})$ is an estimate of $\Pr(Y=1|\pmb{X}=\pmb{x})$.\footnote{Our framework can be easily applied to multi-class classifications, where $f_{\w}(\pmb{X})$ becomes a vector. Because it only complicates the notations without providing  additional insights about our algorithm, we present the method and algorithm in a binary setting.   }  Consider loss function $l:\mathcal{Y}\times \mathbb{R}\to \mathbb{R}$ where $l(Y,f_{\w}(\pmb{X}))$ measures the error of $f_{\pmb{w}}$ in predicting $\pmb{X}$. We denote the expected loss with respect to the joint probability distribution of $(\pmb{X},Y)$ by $L(\pmb{w}) :=\E\{l(Y,f_{\pmb{w}}(\pmb{X}))\}$. Similarly, $L_a(\pmb{w}) :=\E\{l(Y,f_{\pmb{w}}(\pmb{X}))|A=a\}$ denotes the expected loss of the group with sensitive attribute $A=a$. In this work, we assume that $l(y,f_{\w}(\pmb{x}))$ is \textbf{differentiable and strictly convex} in ${\w}$ (e.g., 
binary cross entropy loss).\footnote{We do not consider non-differentiable losses (e.g., zero-one loss) as they have already been extensively studied in the literature, e.g., \cite{hardt2016equality,zafar2017fairness, lohaus2020too}.}

Without fairness consideration, a predictor that simply minimizes the total expected loss, i.e., $\argmin_{\pmb{w}}L(\pmb{w})$, may  be biased against certain groups. To mitigate the risks of unfairness,  we consider \textbf{Equalized Loss (EL)} fairness notion, as formally defined below.

\begin{definition}[$\gamma$-EL \cite{zhang2019group}] \label{def:gammaOAE}
We say $f_{\w}$ satisfies the equalized loss (EL) fairness notion if $L_0({\w}) = L_1({\w})$. Moreover, we say $f_{\w}$ satisfies $\gamma-$EL for some $\gamma>0$ if $-\gamma \leq L_0({\w}) - L_1({\w}) \leq \gamma$.   
\end{definition}
Note that if $l(Y,f_{\pmb{w}}(\pmb{X}))$ is a (strictly) convex function in $\w$, both $L_0(\w)$ and $L_1(\w)$ are also (strictly) convex in ${\w}$. However, $L_0(\w)-L_1(\w)$ is not necessary convex\footnote{As an example, consider two functions $h_0(x) =x^2$ and $h_1(x) = 2\cdot x^2-x$. Although both $h_0$ and $h_1$ are convex, their difference $h_0(x)-h_1(x)$ is not a convex function.  }. As a result, the following optimization problem for finding a fair predictor under $\gamma$-EL is not a convex programming, 
\begin{eqnarray}\label{eq:mainOP}
\min_{\pmb{w}} ~~L(\pmb{w})~\text{ s.t.}~~-\gamma \leq L_0(\pmb{w}) -L_1(\pmb{w}) \leq \gamma. 
\end{eqnarray}
We say a group is \textit{disadvantaged} group if it experiences higher loss than the other group. Before discussing how to find the \textbf{global optimal} solution of the above non-convex optimization problem and train a $\gamma$-EL fair predictor,  we first discuss why $\gamma$-EL is an important fairness notion and why the majority of fair learning algorithms in the literature cannot be used  for finding $\gamma$-EL fair predictors. 

\subsection{Existing Fairness Notions \& Algorithms}
Next, we (mathematically) introduce some of the most commonly used fairness notions and compare them with $\gamma$-EL. We will also discuss why the majority of proposed fair learning algorithms are not properly applicable to EL fairness.

\textbf{Overall Misclassification Rate (OMR)}: It was considered by \cite{zafar2017fairness,zafar2019fairness} for classification problems. Let $\hat{Y} = I(f_{\pmb{w}} (\pmb{X})>0.5)$, where $I(.) \in \{0,1\}$ is an indicator function, and $\hat{Y}=1$ if $f_{\pmb{w}} (\pmb{X})>0.5$. OMR requires $\Pr(\hat{Y}\neq Y|A=0) = \Pr(\hat{Y}\neq Y|A=1),$ 
which is not a convex constraint. As a result, Zafar et al. \yrcite{zafar2017fairness,zafar2019fairness} propose a method to relax this constraint using decision boundary covariances.  
We emphasize that OMR is different from EL fairness, that OMR only equalizes the accuracy of binary predictions across different groups while EL is capable of considering the fairness in estimating probability $\Pr(Y=1|\pmb{X}=\pmb{x})$, e.g., by using binary cross entropy loss function. Note that in many applications such as conversion prediction, click prediction, medical diagnosis, etc., it is critical to find $\Pr(Y=1|\pmb{X}=\pmb{x})$ accurately for different groups besides the final predictions $\hat{Y}$. 
Moreover, unlike EL, OMR is not applicable to regression problems. Therefore, the relaxation method  proposed by \cite{zafar2017fairness,zafar2019fairness} cannot be applied to the EL fairness constraint.

\textbf{Statistical Parity (SP), Equal Opportunity (EO)}: For binary classification, Statistical Parity (SP) \citep{dwork2012fairness} (resp. Equal Opportunity (EO) \citep{hardt2016equality}) requires the positive classification rates (resp. true positive rates) to be equalized across different groups. Formally,  
\begin{eqnarray*}
 \Pr(\hat{Y}=1|A=0) &=& \Pr(\hat{Y}=1|A=1)  \\
 \Pr(\hat{Y}=1|A=0,Y=1) &=& \Pr(\hat{Y}=1|A=1,Y=1)
\end{eqnarray*}
 Both notions can be re-written in the expectation form using an indicator function. Specifically, SP is equivalent to 
    $ \E\{I(f_{\w}(\pmb{X})>0.5)|A=0\} = \E\{I(f_{\w}(\pmb{X})>0.5)|A=1\}$, and EO equals to $ \E\{I(f_{\w}(\pmb{X})>0.5)|A=0,Y=1\} = \E\{I(f_{\w}(\pmb{X})>0.5)|A=1,Y=1\}$.
    Since the indicator function is neither differentiable nor convex,  Donini et al. \yrcite{donini2018empirical} use a linear relaxation of EO as a proxy. \footnote{This linear relaxation is applicable to EL with some modification. We use linear relaxation as one of our baselines.}  However, linear relaxation may negatively affect the fairness of the predictor \citep{lohaus2020too}. To address this issue, Lohaus et al. \yrcite{lohaus2020too} and Wu et al. \yrcite{wu2019convexity} develop convex relaxation techniques for SP and EO fairness criteria by convexifying indicator function $I(.)$.  However, these convex relaxation techniques are not applicable to  EL fairness notion because $l(.,.)$ in our setting is convex, not a zero-one function. FairBatch \cite{roh2020fairbatch} is another algorithm that has been proposed to find a predictor under SP or EO. FairBatch adds a sampling bias in the mini-batch selection. However, the bias in mini-batch sampling distribution leads to a biased estimate of the gradient, and there is \underline{no guarantee} FairBatch finds the global optimum solution. FairBatch can be used to find a \textit{sub-optimal} fair predictor EL fairness notion. We use FairBatch as a baseline. Shen et al. \yrcite{shen2022optimising} propose an algorithm for EO. This algorithm adds a penalty term to the objective function, which is similar to the Penalty Method \cite{ben1997penalty}. We will use the Penalty method as a baseline as well.  
    
    Hardt et al. \yrcite{hardt2016equality} propose a post-processing algorithm that randomly flips the binary predictions to satisfy EO or SP. 
    However, this method does not guarantee finding an optimal classifier \cite{woodworth2017learning}. Agarwal et al. \yrcite{agarwal2018reductions} introduce a reduction approach for SP or EO. However, this method finds a randomized classifier satisfying SP or EO in expectation. In other words, to satisfy {SP}, the reduction approach finds distribution $Q$ over $\mathcal{F}$ such that, 
\begin{eqnarray*}
\textstyle \sum_{f\in \mathcal{F}} Q(f) \E\{l(Y,f(\pmb{X}))|A=0\} \\\textstyle = \sum_{f\in \mathcal{F}} Q(f) \E\{l(Y,f(\pmb{X}))|A=1\}
\end{eqnarray*}
where $Q(f)$ is the probability of selecting model $f$ under distribution $Q$. Obviously, satisfying a fairness constraint in expectation may lead to unfair predictions because $Q$ can still assign a non-zero probability to unfair models.


In summary, maybe some of the approaches used for SP/EO  are applicable to EL fairness notion (e.g., linear relaxation or FairBatch). However, they can only find sub-optimal solutions (see Section \ref{sec:exp} for more details). 

\textbf{Statistical Parity for Regression}: SP can  be adjusted to be suitable for regression. As proposed by \cite{agarwal2019fair}, Statistical Parity for regressor $f_{\w}(.)$ is defined as: 
\begin{equation}\label{eq:SPforReg}
\Pr(f_{\w}(\pmb{X})\leq z|A=a) = \Pr(f_{\w}(\pmb{X})\leq z), \forall z,a. 
\end{equation}
To find a predictor that satisfies constraint \eqref{eq:SPforReg}, Agarwal et al. \yrcite{agarwal2019fair} use the reduction approach as mentioned above. However, this approach only finds a randomized predictor satisfying SP in expectation and cannot be applied to optimization problem \eqref{eq:mainOP}.\footnote{Appendix includes a detailed discussion on why the reduction approach is not appropriate for EL fairness.}


\textbf{Bounded Group Loss (BGL)}:   $\gamma$-BGL  was introduced by \cite{agarwal2019fair} for regression problems. It requires that the loss experienced by each group be bounded by $\gamma$. That is, $L_a(\w)\leq \gamma, \forall a\in\{0,1\}.$   Agarwal et al. \yrcite{agarwal2019fair} use the reduction approach to find a randomized regression model under $\gamma$-BGL. In addition to the reduction method, if  $L(\w)$, $L_0(\w)$, and $L_1(\w)$ are convex in $\w$, then we can directly use convex solvers (e.g., CVXPY \citep{diamond2016cvxpy,agrawal2018rewriting}) to find a $\gamma$-BGL fair predictor. This is because the following  is a convex problem, \begin{equation}\label{eq:BGLopt}
    \min_{\w} L({\w}),~~~ \text{s.t.},~~~ L_a(\w)\leq \gamma, ~\forall a.
    \end{equation}
However, for non-convex  optimization problems such as \eqref{eq:mainOP}, the convex solvers cannot be  used directly. 

We want to emphasize that even though there are already many fairness notions and algorithms in the literature to find a fair predictor, none of the existing algorithms can be used to solve the non-convex optimization \eqref{eq:mainOP} efficiently and  find a global optimal fair predictor under EL notion.

\section{Optimal Model under $\gamma$-EL}\label{sec:EL}
In this section, we consider optimization problem \eqref{eq:mainOP} under {EL} fairness constraint. 
Note that this optimization problem is non-convex and finding the global optimal solution is difficult. We propose an algorithm 
that finds the solution to non-convex optimization \eqref{eq:mainOP}  by solving a sequence of \textit{convex} optimization problems. Before presenting the algorithm, 
we first introduce two assumptions, which will be used when proving the convergence of the proposed algorithm.
\begin{algorithm}[tb]
\caption{Function \texttt{ELminimizer}}
\label{Alg1}
\textbf{Input:} $\pmb{w}_{G_0},\pmb{w}_{G_1},\epsilon,\gamma$\\
\textbf{Parameters:} $\lambda_{start}^{(0)}  = L_0(\pmb{w}_{G_0}), \lambda_{end}^{(0)}  = L_0(\pmb{w}_{G_1}),i=0$ \\
Define $\tilde{L}_1(\pmb{w}) = L_1(\pmb{w}) + \gamma$
\begin{algorithmic}[1]
 \WHILE{$\lambda^{(i)}_{end}-\lambda^{(i)}_{start}>\epsilon$}
\STATE $\lambda^{(i)}_{mid} = (\lambda^{(i)}_{end}+\lambda^{(i)}_{start})/2$
\STATE Solve the following convex optimization problem,\vspace{-0.1cm}
\begin{equation}\label{eq:OptAlg}\pmb{w}_i^* = \arg\min_{\pmb{w}} \tilde{L}_1(\pmb{w})~~\text{s.t.}~~ L_0(\pmb{w})\leq \lambda^{(i)}_{mid}\end{equation}
\STATE $\lambda^{(i)}= \tilde{L}_1(\pmb{w}_i^*) $
\IF{$\lambda^{(i)}\geq \lambda^{(i)}_{mid}$}
\STATE $\lambda^{(i+1)}_{start}=\lambda^{(i)}_{mid}$;~
$\lambda^{(i+1)}_{end}=\lambda^{(i)}_{end}$;
\ELSE
\STATE $\lambda^{(i+1)}_{end}=\lambda^{(i)}_{mid}$;~
$\lambda^{(i+1)}_{start}=\lambda^{(i)}_{start}$;
\STATE $i = i+1$;
\ENDIF
\ENDWHILE
\end{algorithmic}
\textbf{Output:} $\pmb{w}_i^*$
\end{algorithm}

\begin{assumption}\label{assump:1} Expected losses
$L_0(\pmb{w})$, $L_1(\pmb{w})$, and $L(\pmb{w})$ are strictly convex and differentiable in $\pmb{w}$. Moreover, each of them has a unique minimizer. 
\end{assumption}

Let $\pmb{w}_{G_a}$ be the optimal weight vector minimizing the loss associated with group $A=a$. That is, \begin{equation}\label{eq:UnconstratintGroupBased}
\pmb{w}_{G_a} = \arg\min_{\pmb{w}} L_a(\pmb{w}).
\end{equation}
Since problem  \eqref{eq:UnconstratintGroupBased} is an \textit{unconstrained}, \textit{convex} optimization problem,   $\pmb{w}_{G_a}$ can be found efficiently by common convex solvers. We make the following assumption about $\pmb{w}_{G_a}$. 
\begin{assumption}\label{assump:2}

We assume the following holds, 
$$L_{0}(\pmb{w}_{G_{0}})\leq L_{1}(\pmb{w}_{G_{0}}) \text{ and } L_{1}(\pmb{w}_{G_{1}})\leq L_{0}(\pmb{w}_{G_{1}}). $$ 
\end{assumption}

\begin{algorithm}[tb]
\caption{Solving Optimization (\ref{eq:mainOP})}\label{Alg2}
\textbf{Input:} $\pmb{w}_{G_0}$, $\pmb{w}_{G_1}$,$\epsilon$,$\gamma$
\begin{algorithmic}[1]
\STATE $\pmb{w}_{\gamma} = \texttt{ELminimizer}(\pmb{w}_{G_0}, \pmb{w}_{G_1},\epsilon,\gamma )$
\STATE$\pmb{w}_{-\gamma} = \texttt{ELminimizer}(\pmb{w}_{G_0}, \pmb{w}_{G_1},\epsilon,-\gamma )$
\IF{$L(\pmb{w}_{\gamma}) \leq  L(\pmb{w}_{-\gamma})$}
\STATE$\pmb{w}^* = \pmb{w}_{\gamma}$
\ELSE
\STATE $\pmb{w}^* = \pmb{w}_{-\gamma}$
\ENDIF
\end{algorithmic}
\textbf{Output:} $\pmb{w}^*$
\end{algorithm}

Assumption \ref{assump:2} implies that when a group experiences its lowest possible loss, this group is not the disadvantaged group.
Under Assumptions \ref{assump:1} and \ref{assump:2}, the optimal 0-{EL} fair predictor can be easily found using our proposed algorithm (i.e., function $\texttt{ELminimizer} (\pmb{w}_{G_0},\pmb{w}_{G_1},\epsilon,\gamma)$ with  $\gamma=0$); the complete procedure is shown in Algorithm \ref{Alg1}, in which
parameter $\epsilon>0$ specifies the stopping criterion: as $\epsilon\to 0$, the output approaches to the global optimal solution.

Intuitively, Algorithm 1 solves non-convex optimization \eqref{eq:mainOP} by solving a sequence of convex and constrained optimizations. When $\gamma>0$ (i.e., relaxed fairness), the optimal $\gamma$-{EL} fair predictor can be found with Algorithm \ref{Alg2} which calls function $\texttt{ELminimizer}$ twice. The convergence of Algorithm \ref{Alg1} for finding the optimal 0-{EL} fair solution, and the convergence of Algorithm \ref{Alg2} for finding the optimal $\gamma$-{EL} fair solution are stated in the following theorems.

\begin{theorem}[Convergence of Algorithm \ref{Alg1} when $\gamma=0$]\label{theo:main}
Let $\{\lambda^{(i)}_{mid}| i=0,1,2,\ldots\}$ and $\{\pmb{{w}}_{i}^*| i=0,1,2,\ldots\}$ be two sequences generated by Algorithm \ref{Alg1} when $\gamma=\epsilon = 0$, i.e., $\texttt{ELminimizer} (\pmb{w}_{G_0},\pmb{w}_{G_1},0,0).$ Under Assumptions \ref{assump:1} and \ref{assump:2}, we have,
$$\lim_{i\to \infty}  \pmb{{w}}_{i}^* = \pmb{{w}}^* \text{ and } \lim_{i\to \infty}  \lambda^{(i)}_{mid} = \E\{l(Y,f_{\pmb{w}^*}(\pmb{X}))\}$$
where $\pmb{w}^*$ is the global optimal solution to \eqref{eq:mainOP}.
\end{theorem}
Theorem \ref{theo:main} implies that when $\gamma = \epsilon = 0$ and $i$ goes to infinity, the solution to convex problem \eqref{eq:OptAlg} is the same as the solution to \eqref{eq:mainOP}.
\begin{theorem}[Convergence of Algorithm \ref{Alg2}]\label{theo:gammaEL}
Assume that $L_0(\pmb{w}_{G_0}) -L_1(\pmb{w}_{G_0}) < -\gamma $ and $L_0(\pmb{w}_{G_1}) -L_1(\pmb{w}_{G_1}) > \gamma $. If $\w_O$ does not satisfy the $\gamma$-EL constraint, then, as $\epsilon \to 0$, the output of Algorithm \ref{Alg2} goes to the optimal $\gamma$-{EL} fair solution (i.e., solution to \eqref{eq:mainOP}).
\end{theorem}

\textbf{Complexity Analysis.} The \texttt{While} loop in Algorithm \ref{Alg1} is executed for $\mathcal{O}(\log(1/\epsilon))$ times. Therefore, Algorithm \ref{Alg1} needs to solve a constrained convex optimization problem for $\mathcal{O}(\log(1/\epsilon))$ times. Note that constrained convex optimization problems can be efficiently solved via sub-gradient methods \citep{nedic2009subgradient}, brier methods \citep{wright2001convergence}, stochastic gradient descent with one projection \citep{mahdavi2012stochastic}, interior point methods \citep{nemirovski2004interior}, etc.  For instance, \cite{nemirovski2004interior} shows that several convex optimization problems can be solved in polynomial time. Therefore, the time complexity of Algorithm \ref{Alg1} depends on the convex solver. If the time complexity of solving \eqref{eq:OptAlg} is $\mathcal{O}(p(d_{\w}))$, then the overall time complexity of Algorithm \ref{Alg1} is $\mathcal{O}(p(d_{\w})\log(1/\epsilon))$.   

\textbf{Regularization.}
So far we have considered a supervised learning model without regularization. Next, we explain how Algorithm \ref{Alg2} can be applied to a regularized problem. 
Consider the following optimization problem, 
\begin{eqnarray}
\min_{\w} && \Pr(A=0) L_0(\w) + \Pr(A=1) L_1(\w)+R(\w),\nonumber\\
\text{s.t.},&& |L_0(\w) - L_1(\w)|<\gamma. \label{eq:regularized}
\end{eqnarray}

where $R(\w)$ is a regularizer function. In this case, we can re-write the optimization problem as follows,
\begin{eqnarray*}
\min_{\w} &&\Pr(A=0) \big(L_0(\w) +R(\w)\big)\\&&+ \Pr(A=1) \big(L_1(\w)+R(\w)\big),\\
\text{s.t.}, &&|\big(L_0(\w)+R(\w)\big) - \big(L_1(\w)+R(\w)\big)|<\gamma.
\end{eqnarray*}

If we define $\bar{L}_a({\w}) :=L_a(\w)+R(\w)$ and 
$\bar{L}(\w) := \Pr(A=0)\bar{L}_0(\w) + \Pr(A=1)\bar{L}_1(\w),$ 
then  problem \eqref{eq:regularized} can be written in the form of problem \eqref{eq:mainOP} using $(\bar{L}_0({\w}),\bar{L}_1({\w}),\bar{L}({\w}))$ and solved by Algorithm \ref{Alg2}.

\section{Sub-optimal Model under $\gamma$-{EL}}\label{sec:gamma_EL}
In Section \ref{sec:EL}, we have shown that non-convex optimization problem  \eqref{eq:mainOP} can be reduced to a sequence of convex constrained optimizations \eqref{eq:OptAlg}, and based on this we proposed Algorithm \ref{Alg2} that finds the optimal $\gamma$-{EL} fair predictor. However, the proposed algorithm still requires solving a convex constrained optimization in each iteration. In this section, we propose another algorithm that finds a \textit{sub-optimal} solution to optimization \eqref{eq:mainOP} without solving constrained optimization in each iteration.   
The algorithm consists of two phases: (1) finding two weight vectors by solving two \textit{unconstrained} convex optimization problems; (2) generating a new weight vector satisfying $\gamma$-{EL} using the two weight vectors found in the first phase. 

\textbf{Phase 1: Unconstrained optimization.}
We ignore {EL} fairness  and  solve the following unconstrained problem, 
\begin{equation}\label{eq:Unconstratint1}
\pmb{w}_O = \arg\min_{\pmb{w}} L(\w)
\end{equation}
Because $L(\w) $ is strictly convex in $\w$, the above optimization problem can be solved efficiently using convex solvers. Predictor $f_{\pmb{w}_O}$ is the optimal predictor without fairness constraint, and $L(\w_O)$ is the smallest overall expected loss that is attainable. Let $
\hat{a} =  \arg\max_{a\in \{0,1\}} L_a(\w_O)$, i.e., group $\hat{a}$ is  disadvantaged under predictor $f_{\pmb{w}_O}$. Then, for the disadvantaged group $\hat{a}$, we find $\w_{G_{\hat{a}}}$ by  optimization  \eqref{eq:UnconstratintGroupBased}.  

\textbf{Phase 2: Binary search to find the fair predictor.} For $\beta \in [0,1]$, we define the following two functions,
\begin{eqnarray*}
g(\beta) &:=& L_{\hat{a}}\big((1-\beta) \pmb{w}_O + \beta \w_{G_{\hat{a}}}\big)\\ &&- L_{1-\hat{a}}\big((1-\beta) \pmb{w}_O + \beta \w_{G_{\hat{a}}}\big);\\
h(\beta) &:=& L\big((1-\beta) \pmb{w}_O + \beta \w_{G_{\hat{a}}} \big), 
\end{eqnarray*} where function $g(\beta)$ can be interpreted as the loss disparity between two demographic groups under predictor $f_{(1-\beta) \pmb{w}_O + \beta \w_{G_{\hat{a}}}}$, and $h(\beta)$ is the corresponding overall expected loss.
Some properties of functions $g(.)$ and $h(.)$ are summarized in the following theorem.

\begin{theorem}\label{Theo:3}Under Assumptions \ref{assump:1} and \ref{assump:2},
$\newline$1. There exists $\beta_0 \in [0,1]$ such that $g(\beta_0)=0$;
$\newline$
 2. $h(\beta)$ is strictly increasing in $\beta \in [0,1]$; 
 $\newline$
 3. $g(\beta)$ is strictly decreasing in $\beta \in [0,1]$.
\end{theorem}
Theorem \ref{Theo:3} implies that in a $d_{\w}$-dimensional space if we start from $\pmb{w}_O$ and move toward $\pmb{w}_{G_{\hat{a}}}$ along a straight line, the overall loss increases and the disparity between two groups decreases until we reach $(1-\beta_0)\pmb{w}_O + \beta_0\pmb{w}_{G_{\hat{a}}}$, at which 0-{EL} fairness is satisfied.
Note that $\beta_0$ is the unique root of $g$. Since $g(\beta)$ is a strictly decreasing function, $\beta_0$ can be found using \textit{binary search}. 

For the approximate $\gamma$-{EL} fairness, there are multiple values of $\beta$ such that $(1-\beta)\pmb{w}_O + \beta\pmb{w}_{G_{\hat{a}}}$ satisfies $\gamma$-{EL}. Since $h(\beta)$ is strictly increasing in $\beta$, among all  $\beta$ that satisfy $\gamma$-{EL} fairness, we would choose the smallest one. The method for finding a sub-optimal solution to optimization 
\eqref{eq:mainOP} is described in Algorithm \ref{Alg3}.  
\begin{algorithm}[tb]
\caption{Sub-optimal solution to optimization (\ref{eq:mainOP})}\label{Alg3}
\textbf{Input:} $\pmb{w}_{G_{\hat{a}}}$, $\pmb{w}_{O}$, $\epsilon$, $\gamma$\\ 
\textbf{Initialization:} $g_\gamma(\beta) = g(\beta)-\gamma$, $i=0$,  $\beta_{start}^{(0)}=0$, $\beta_{end}^{(0)}=1$
\begin{algorithmic}[1]
\IF{$g_\gamma(0)\leq 0$}
\STATE $\underline{\w} = \w_O$, and go to line {13};
\ENDIF
\WHILE{$\beta_{end}^{(i)} - \beta_{start}^{(i)} > \epsilon$}
\STATE $\beta^{(i)}_{mid} = (\beta_{start}^{(i)} + \beta_{end}^{(i)})/2$;
\IF{$g_{\gamma}(\beta_{mid}^{(i)})\geq 0 $}
\STATE$\beta_{start}^{(i+1)} = \beta_{mid}^{(i)},$
$\beta_{end}^{(i+1)} = \beta_{end}^{(i)}$;
\ELSE
\STATE$\beta_{start}^{(i+1)} = \beta_{start}^{(i)},$ 
$\beta_{end}^{(i+1)} = \beta_{mid}^{(i)}$;
\ENDIF
\ENDWHILE
\STATE $\underline{\w} = (1-\beta_{mid}^{(i)})\w_O + \beta_{mid}^{(i)}\w_{G_{\hat{a}}}$;
\STATE\textbf{Output:} {$\underline{\pmb{w}}$}

\end{algorithmic}
\end{algorithm}
Note that  \texttt{while} loop in Algorithm \ref{Alg3} is repeated for $\mathcal{O}(\log(1/\epsilon))$ times. Since the time complexity of operations (i.e., evaluating $g_{\gamma}(\beta_{mid}^{(i)})$) in each iteration is $\mathcal{O}(d_{\w})$ ,  the total time complexity of Algorithm \ref{Alg3} is $\mathcal{O}(d_{\w}\log(1/\epsilon))$. We can formally prove that the output returned by Algorithm \ref{Alg3} satisfies $\gamma$-{EL}  constraint. 
\begin{theorem}\label{theo:sub}
Assume that Assumptions \ref{assump:1} and \ref{assump:2} hold, and let $g_\gamma(\beta) = g(\beta)-\gamma$. If $g_{\gamma}(0)\leq 0$, then $\w_O$ satisfies the $\gamma$-{EL} fairness; if $g_{\gamma}(0)> 0$, then  $\lim_{i\to \infty} \beta_{mid}^{(i)} = \beta_{mid}^{(\infty)}$ exists, and  $(1-\beta_{mid}^{(\infty)})\w_O + \beta_{mid}^{(\infty)}\w_{G_{\hat{a}}}$ satisfies the $\gamma$-{EL} fairness constraint.
\end{theorem}
Note that since $h(\beta)$ is increasing in $\beta$, we only need to find the smallest possible $\beta$ such that $(1-\beta)\w_O + \beta\w_{G_{\hat{a}}}$ satisfies $\gamma$-{EL}, which is $\beta_{mid}^{(\infty)}$ in Theorem \ref{theo:sub}.  
Since Algorithm \ref{Alg3} finds a sub-optimal solution, it is important to investigate the performance of this sub-optimal fair predictor, especially in the worst case scenario. The following theorem finds an upper bound of the expected loss of  $f_{\pmb{\underline{w}}}$, where $\pmb{\underline{w}}$ is the output of Algorithm \ref{Alg3}. 
\begin{theorem}\label{theo:sub_perf}
Under Assumptions  \ref{assump:1} and \ref{assump:2}, we have the following: $L(\pmb{\underline{w}}) \leq \max_{a\in \{0,1\}} L_a(\w_O).$ That is, the expected loss of $f_{\pmb{\underline{w}}}$ is not worse than the  loss of the disadvantaged group under predictor $f_{\pmb{w}_O}$. 
\end{theorem}

\textbf{Learning with Finite Samples.}\label{sec:Gen}
So far we proposed algorithms for  solving optimization \eqref{eq:mainOP}. In practice, the joint probability distribution of $(\pmb{X},A,Y)$ is unknown and the expected loss needs to be estimated using the empirical loss. Specifically, given $n$ i.i.d. samples $\{(\pmb{X}_i,A_i,Y_i)\}_{i=1}^n$ and a predictor $f_{\w}$, the empirical losses of the entire population and each group are defined as follows,
\begin{eqnarray*}
\textstyle\hat{L}(\pmb{w}) &=& \textstyle \frac{1}{n}\sum_{i=1}^n l(Y_i,f_{\pmb{w}}(\pmb{X}_i)),\\ \textstyle\hat{L}_a(\pmb{w}) &=& \textstyle \frac{1}{n_a}\sum_{i: A_i=a} l(Y_i,f_{\pmb{w}}(\pmb{X}_i)),
\end{eqnarray*}
where $n_a = |\{i|A_i=a\}|$. Because $\gamma$-{EL} fairness constraint is defined in terms of expected loss, the optimization problem of finding an optimal $\gamma$-{EL} fair predictor using empirical losses is as follows,
\begin{eqnarray}\label{eq:approx}
    \pmb{\hat{w}} = \arg\min_{\pmb{w}}\hat{L}(\pmb{w}),~~~ \text{s.t.}~~~ |\hat{L}_0(\pmb{w})-\hat{L}_1(\pmb{w})| \leq \hat{\gamma}.
\end{eqnarray}
In this section, we aim to investigate how to determine $\hat{\gamma}$ so that with high probability, the predictor found by solving problem \eqref{eq:approx} satisfies $\gamma$-{EL} fairness, and meanwhile  $\pmb{\hat{w}}$ is a good estimate of the solution $\pmb{w}^*$  to optimization \eqref{eq:mainOP}. We aim to show that we can set $\hat{\gamma}=\gamma$ if the number of samples is sufficiently large. To understand the relation between \eqref{eq:approx} and \eqref{eq:mainOP}, we follow the general sample complexity analysis found in \cite{donini2018empirical} and show their sample complexity analysis is applicable to EL. To proceed, we make the assumption used in \cite{donini2018empirical}. 
\begin{assumption}\label{assump:learnable}
With probability $1-\delta$, following holds:
$$\textstyle \sup_{f_{\pmb{w}}\in \mathcal{F}} |L(\pmb{w}) - \hat{L}(\pmb{w})| \leq B(\delta,n,\mathcal{F}), $$
where $B(\delta,n,\mathcal{F})$ is a bound that goes to zero as $n\to+\infty$.
\end{assumption}
Note that according to \cite{shalev2014understanding}, if the class $\mathcal{F}$ is learnable with respect to loss function $l(.,.)$, then always there exists such a bound $B(\delta,n,\mathcal{F})$ that goes to zero as $n$ goes to infinity.\footnote{As an example, if $\mathcal{F}$ is a compact subset of linear predictors in Reproducing Kernel Hilbert Space (RKHS) and loss $l(y,f(x))$ is Lipschitz in $f(x)$ (second argument), then Assumption \ref{assump:learnable} can be satisfied \cite{bartlett2002rademacher}.   Vast majority of linear predictors such as support vector machine and  logistic regression can be defined in RKHS.} 
\begin{theorem}\label{theo:learnable}
Let $\mathcal{F}$ be a set of learnable functions, and let $\hat{\pmb{w}}$ and $\pmb{w}^*$ be the solutions to \eqref{eq:approx} and \eqref{eq:mainOP} respectively, with $\linebreak\hat{\gamma} = \gamma + \sum_{a\in\{0,1\}}B(\delta,n_a,\mathcal{F}).$ Then, with probability at least $1-6\delta$, the followings hold, 
\begin{eqnarray*}
L(\pmb{\hat{w}}) - L(\pmb{w}^*)  &\leq& 2 B(\delta,n,\mathcal{F})~\text{ and }\\  |{L}_0(\pmb{\hat{w}})-{L}_1(\pmb{\hat{w}})| &\leq& \gamma + 2B(\delta,n_0,\mathcal{F}) + 2B(\delta,n_1,\mathcal{F}).
\end{eqnarray*}
\end{theorem}
Theorem \ref{theo:learnable} shows that as $n_0$, $n_1$ go to infinity, $\hat{\gamma} \to \gamma$, and both empirical loss and expected loss satisfy $\gamma$-{EL}. In addition, as $n$ goes to infinity, the expected loss at $\hat{\w}$ goes to the minimum possible expected loss. Therefore, solving \eqref{eq:approx} using empirical loss is equivalent to solving \eqref{eq:mainOP} if the number of data points from each group  is sufficiently large. 
\section{Beyond Linear Models}
So far, we have assumed that the loss function is strictly convex. This assumption is mainly  valid for training linear models (e.g., Ridge regression, regularized logistic regression). 
However, it is known that training deep models lead to minimizing  non-convex objective functions. To train a deep model under the equalized loss fairness notion, we can take advantage of Algorithm 2 for fine-tuning under EL as long as the objective function is convex with respect to the parameters of the output layer.\footnote{In classification or regression problems with l2 regularizer, the objective function is strictly convex with respect to the parameters of the output layer. This is true regardless of the network structure before the output layer. }  
To clarify how Algorithm 2 can be used for deep models, for simplicity, consider a neural network with one hidden layer for regression. Let $W$ be an $m$ by $d$ matrix ($d$ is the size of feature vector $\pmb{X}$ and $m$ is the number of neurons in the first layer) denoting the parameters of the first layer of the Neural Network, and $\pmb{w}$ be a vector corresponding to the output layer. 
To find a neural network satisfying the equalized loss fairness notion, first, we train the network without any fairness constraint  using common gradient descent algorithms (e.g., stochastic gradient descent). Let $\tilde{W}$ and $\tilde{\pmb{w}}$ denote the network parameters after training the network without  fairness constraint. Now we can take advantage of Algorithm 2 to fine-tune the parameters of the output layer under the equalized loss fairness notion. 
Let us define $\tilde{\pmb{X}}:= [1,\tilde{W}\cdot \pmb{X}]^T$. The problem for fine-tuning the output layer can be written as follows, 
\begin{eqnarray}\label{eq:fineTune}
\pmb{w}^* &=& \arg\min_{\pmb{w}}\E\{l(Y,\pmb{w}^T\tilde{\pmb{X}})\},\\
\text{s.t.},&& \hspace{-0.72cm}\left|\E\{l(Y,\pmb{w}^T\tilde{\pmb{X}})|A=0\} - \E\{l(Y,\pmb{w}^T\tilde{\pmb{X}})|A=1\}\right|\leq \gamma.  \nonumber
\end{eqnarray}

The objective function of the above optimization problem is strictly convex, and the optimization problem can be solved using Algorithm 2. After solving the above problem, $[\tilde{W},\pmb{w}^*]$ will be the final parameters of the neural network model satisfying the equalized loss fairness notion. Note that a similar optimization problem can be written for  fine-tuning any deep model with classification/regression task. 

\section{Experiments}\label{sec:exp}We conduct experiments on two real-world datasets to evaluate the performance of the  proposed algorithm. In our experiments, we used a system with the following configurations: 24 GB of RAM, 2 cores of P100-16GB GPU, and 2 cores of Intel Xeon CPU@2.3 GHz processor. More information about the experiments and the instructions on reproducing the empirical results are provided in Appendix. The codes are available at \texttt{\url{https://github.com/KhaliliMahdi/Loss_Balancing_ICML2023}}.

\textbf{Baselines.}
As discussed in Section \ref{sec:formulation}, not all the fair learning algorithms are applicable to EL fairness. The followings are three baselines that are applicable to EL fairness. 

\textbf{Penalty Method (PM):} The penalty method \cite{ben1997penalty}  finds a fair predictor under $\gamma$-EL fairness constraint by solving  the following  problem,
{\small
\begin{eqnarray}\label{eq:penalty}
\min_{\w} \hat{L}(\w) + t \cdot \max\{0, |\hat{L}_0(\w)-\hat{L}_1(\w)|- \gamma\}^{2} +R(\w),
\end{eqnarray}}
where $t$ is the penalty parameter, and $R(\w)=0.002\cdot||\w||_2^2$ is the regularizer. The above optimization problem cannot be solved with a convex solver because it is not generally convex. We solve problem \eqref{eq:penalty} using Adam gradient descent \citep{kingma2014adam} with a learning rate of $0.005$. We use the default parameters of Adam optimization in Pytorch.  
We set the penalty parameter $t=0.1$ and increase this penalty coefficient by a factor of $2$ every $100$ iteration. 

 \textbf{Linear Relaxation (LinRe):} Inspired by \cite{donini2018empirical}, for the linear regression, we relax the EL constraint as $-\gamma\leq \frac{1}{n_0}\sum_{i: A_i=a} (Y_i-\pmb{w}^T\pmb{X}_i) - \frac{1}{n_1}\sum_{i: A_i=1} (Y_i-\pmb{w}^T\pmb{X}_i) \leq \gamma$. For the logistic regression, we relax the constraint as $-\gamma\leq \frac{1}{n_0}\sum_{i: A_i=a} (Y_i-0.5)\cdot(\pmb{w}^T\pmb{X}_i) - \frac{1}{n_1}\sum_{i: A_i=1} (Y_i-0.5)\cdot (\pmb{w}^T\pmb{X}_i) \leq \gamma$. Note that the sign of $(Y_i-0.5)\cdot(\pmb{w}^T\pmb{X}_i)$ determines whether the binary classifier makes a correct prediction or not. 
 
 \textbf{FairBatch }\cite{roh2020fairbatch}: This method was originally proposed for equal opportunity, statistical parity, and equalized odds. With some modifications (see the appendix for more details), this can be applied to EL fairness.  This algorithm measures the loss of each group in each epoch and changes the Minibatch sampling distribution in favor of the group with a higher empirical loss. When implementing FairBatch, we use Adam optimization with default parameters, a learning rate of $0.005$, and a batch size of $100$. 

\textbf{Linear Regression and Law School Admission Dataset.}
In the first experiment, we use the law school admission dataset, which includes the information of 21,790 law students studying in 163 different law schools across the United States \citep{wightman1998lsac}. This dataset contains  entrance exam scores (LSAT),
grade-point average (GPA) prior to law school, and the first year average grade (FYA). Our goal is to train a $\gamma$-{EL} fair regularized linear regression model to estimate the FYA of students given their LSAT and GPA. In this study, we consider Black and White Demographic groups. In this dataset, $18285$ data points belong to White students, and $1282$  data points are from Black students. We randomly split the dataset into training and test datasets  (70\% for training and 30\% for testing), and conduct five independent runs of the experiment. A fair predictor is found by solving the following optimization problem,
\begin{eqnarray}\label{eq:exp}
\min_{\w} \hat{L}(\w) + 0.002\cdot||\w||_2^2~~\text{s.t.}, |\hat{L}_0(\w)-\hat{L}_1(\w)|\leq \gamma,
\end{eqnarray}
with $\hat{L}$ and $\hat{L}_a$ being the overall and the group specific empirical MSE, respectively. Note that $0.002\cdot||\w||_2^2$ is the regularizer. We use Algorithm 2 and Algorithm 3 with $\epsilon = 0.01$ to find the optimal linear regression model under {EL} and adopt CVXPY python library \citep{diamond2016cvxpy,agrawal2018rewriting} as the convex optimization solver in \texttt{ELminimizer} algorithm. 
\begin{figure}[t]
	\begin{minipage}{0.24\textwidth}
	\centering
		\includegraphics[width=1\linewidth]{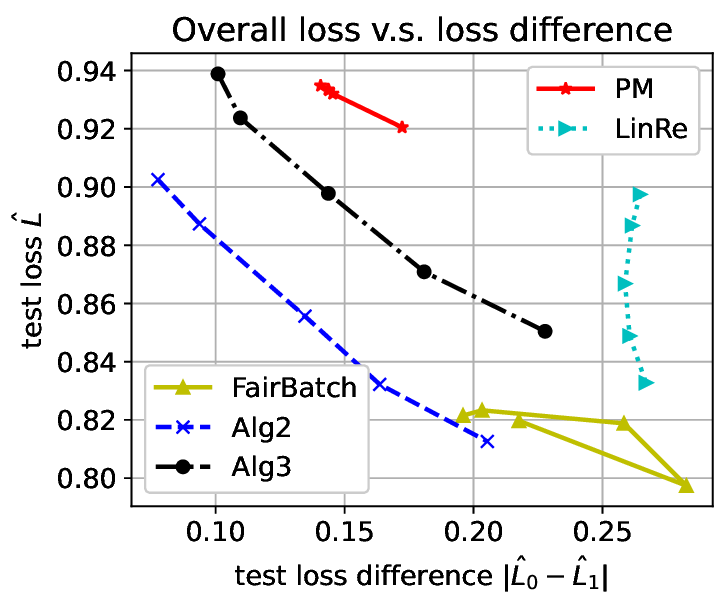}
		\vspace{0.0cm}	\caption{Trade-off between overall MSE and unfairness. A lower curve implies a better trade-off. }
		\label{fig:MSE}
	\end{minipage}~~~
	\begin{minipage}{.24\textwidth}
	\centering
		\includegraphics[width=1\linewidth]{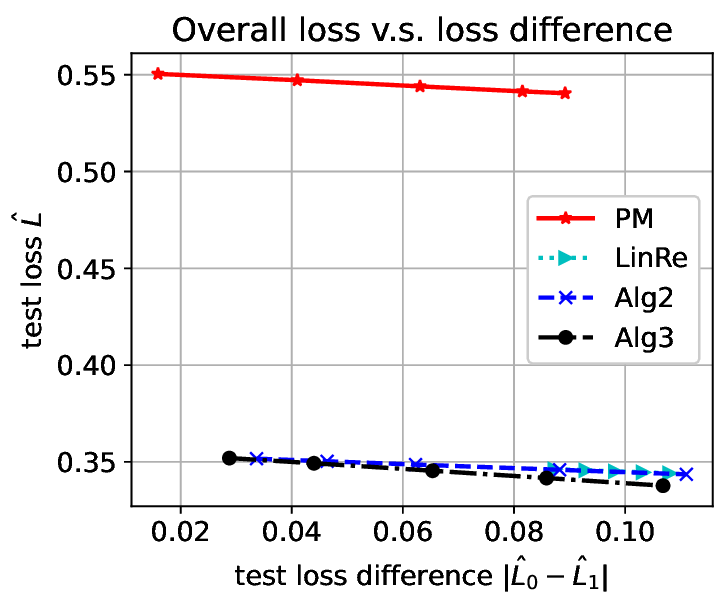}
		\vspace{0.0cm}
		\caption{ Trade-off between overall BCE and unfairness. A lower curve implies a better trade-off.} 
		\label{fig:BCE}
	\end{minipage}
\vspace{-0.3cm}
\end{figure}

\begin{table}[t!]
\caption{Linear regression model under {EL} fairness. The loss function in this example is the mean squared error loss.  } 
\centering
\resizebox{\linewidth}{!}{
\begin{tabular}{cccc} 
\toprule
& &$\gamma =0$ &  $\gamma = 0.1$ \\
\midrule
 &test loss 	&  $0.9246\pm 0.0083$ &  $0.9332\pm 0.0101$  \\
\rot{\rlap{~{PM}}} &test $|\hat{L}_0 - \hat{L}_1| $ 	&$0.1620\pm 0.0802$ & $0.1438\pm 0.0914$\\
\midrule
&  test loss  & $0.9086\pm 0.0190$ & $0.8668\pm  0.0164$ \\
\rot{\rlap{{\small{LinRe}}}} &test $|\hat{L}_0 - \hat{L}_1| $ & $0.2687\pm 0.0588$ &$0.2587\pm  0.0704$ \\
\midrule
&  test loss  & $0.8119\pm 0.0316$ & $0.8610\pm  0.0884$ \\
\rot{\rlap{~{\small{Fair}}}}
\rot{\rlap{~{\small{Batch}}}} &test $|\hat{L}_0 - \hat{L}_1| $ & $0.2862\pm 0.1933$ &$0.2708\pm  0.1526$ \\
\midrule
&test loss	& ${0.9186}\pm {0.0179}$ &  ${0.8556}\pm {0.0217}$  \\
\rot{\rlap{{ours}}}
\rot{\rlap{{Alg 2}}}&test $|\hat{L}_0-\hat{L}_1|$& $0.0699\pm 0.0469$ & $0.1346\pm 0.0749$ \\\midrule
 &test loss &$0.9522\pm 0.0209$&$0.8977\pm 0.0223$\\
 \rot{\rlap{{ours}}}
 \rot{\rlap{{Alg 3}}}&test $|\hat{L}_0-\hat{L}_1|$ &$0.0930\pm 0.0475$&$0.1437\pm0.0907$\\
\bottomrule 
\end{tabular}}
\label{Table2}
\end{table}
Table \ref{Table2} illustrates the means and  standard deviations of empirical loss and the loss difference between Black and White students. The first row specifies desired fairness level ($\gamma=0$ and $\gamma = 0.1$) used as the input to each algorithm. Based on Table \ref{Table2}, when desired fairness level is $\gamma=0$, the model fairness level trained by LinRe and FairBatch method is far from $\gamma = 0$. We also realized that the performance of FairBatch highly depends on the random seed. As a result, the fairness level of the model trained by FairBatch has  a high variance (0.1933 in this example) in these five independent runs of the experiment, and in some of these runs, it can achieve desired fairness level. This is because the FairBatch algorithm does not come with any performance guarantee, and as stated in \cite{roh2020fairbatch}, FairBatch calculates a biased estimate of the gradient in each epoch, and the mini-batch sampling distribution keeps changing from one epoch to another epoch. We observed that FairBatch has better performance with a non-linear model (see Table \ref{Table3}).  Both Algorithms 2 and  3 can achieve a fairness level close to $\gamma=0 $. However, Algorithm 3 finds a sub-optimal solution and achieves higher MSE compared to Algorithm 2. For $\gamma=0.1$, in addition to Algorithms 2 and   3, the penalty method also achieves a fairness level close to desired fairness level $\gamma=0.1$ (i.e., $|\hat{L}_1-\hat{L}_0| = 0.0892$). Algorithm 2 still achieves the lowest MSE compared to Algorithm 3 and the penalty method. The model trained by FairBatch also suffers from high variance in the fairness level.  We want to emphasize that even though Algorithm 3 has a higher MSE compared to Algorithm 2, it is much faster, as stated in Section \ref{sec:EL}.

We also investigate the trade-off between fairness and overall loss under different algorithms. Figure \ref{fig:MSE} illustrates the MSE loss as a function of the loss difference between Black and White students. Specifically, we run Algorithm 2, Algorithm 3, and the baselines under different values of $\gamma = [0.0250,0.05,0.1,0.15,0.2]$. For each $\gamma$, we repeat the experiment five times and calculate the average MSE and average MSE difference over these five runs using the test dataset. Figure \ref{fig:MSE}  shows the penalty method, linear relaxation, and FairBatch are not sensitive to input $\gamma$.  However, Algorithm 2 and Algorithm 3 are sensitive to $\gamma$; As $\gamma$ increases, $|\hat{L}_0(\pmb{w}^*)-\hat{L}_1(\pmb{w}^*)|$ increases and  MSE decreases.  

\textbf{Logistic Regression and Adult Income Dataset.}
We consider the adult income dataset  containing the information of 48,842 individuals \cite{kohavi1996scaling}. Each data point consists of 14 features, including age, education,  race, etc. In this study, we consider race (White or Black) as the sensitive attribute and denote the White demographic group  by $A=0$ and the Black group by $A=1$.  We first pre-process the dataset by removing the data points with a  missing value or with a race other than Black and White; this results in 41,961 data points. Among these data points, 4585 belong to the Black  group. 
For each data point, we convert all the categorical features to one-hot vectors with $110$ dimension and 
randomly split the dataset into training and test data sets (70\% of the dataset is used for training). The goal is to predict whether the income of an individual is above $\$50K$ using a $\gamma$-EL fair logistic regression model. In this experiment, we solve optimization problem \eqref{eq:exp},
with $\hat{L}$ and $\hat{L}_a$ being the overall and the group specific empirical average of binary cross entropy (BCE) loss, respectively. 
\begin{table}[t!]
\caption{Logistic Regression model under {EL} fairness. The loss function in this example is binary cross entropy loss.  } 
\centering
\resizebox{\linewidth}{!}{
\begin{tabular}{cccc} 
\toprule
& &$\gamma =0$ &  $\gamma = 0.1$ \\
\midrule
&test loss 	&  $0.5594\pm 0.0101$ &  $0.5404\pm 0.0046$   \\
\rot{\rlap{~{PM}}} &test $|\hat{L}_0 - \hat{L}_1| $ 	&$0.0091\pm 0.0067$  & $0.0892\pm 0.0378$ \\
\midrule
&  test loss  & $0.3468\pm 0.0013$ & $0.3441\pm  0.0012$ \\
\rot{\rlap{\small{LinRe}}} &test $|\hat{L}_0 - \hat{L}_1| $ & $0.0815\pm 0.0098$ &$0.1080\pm  0.0098$ \\
\midrule
&test loss	& $1.5716\pm 0.8071$ &   $1.2116\pm 0.8819$ \\
\rot{\rlap{~{\small{Fair}}}}
\rot{\rlap{~{\small{Batch}}}} &test $|\hat{L}_0 - \hat{L}_1| $ & $0.6191\pm 0.5459$ &$ 0.3815\pm  0.3470$ \\
\midrule
&test loss 	& ${0.3516}\pm {0.0015}$ &  $0.3435\pm 0.0012$   \\
\rot{\rlap{\small{Ours}}}
\rot{\rlap{\small{Alg2}}}&test $|\hat{L}_0-\hat{L}_1|$& $0.0336\pm 0.0075$ & $0.1110\pm 0.0140$ \\\midrule
 &test loss &$0.3521\pm 0.0015$& ${0.3377}\pm 0.0015$\\
 \rot{\rlap{\small{Ours}}}
 \rot{\rlap{\small{Alg3}}}&test $|\hat{L}_0-\hat{L}_1|$ &$0.0278\pm0.0075$&$0.1068\pm0.0138$\\
\bottomrule 
\end{tabular}}
\label{Table}
\end{table}
The comparison of Algorithm 2, Algorithm 3, and the baselines is shown in Table \ref{Table}, where we conduct five independent runs of experiments, and calculate the mean and standard deviation of overall loss and the loss difference between two demographic groups. The first row in this table shows the value of $\gamma$ used as an input to the algorithms. The results show that linear relaxation, algorithm 2 and Algorithm 3 have very similar performances. All of these three algorithms are able to satisfy the $\gamma$-EL with small test loss. Similar to Table \ref{Table2}, we observe the high variance in the performance of FairBatch, which highly depends on the random seed.    

In Figure \ref{fig:BCE}, we compare the performance-fairness trade-off. We focus on binary cross entropy on the test dataset. To generate this figure, we run Algorithm 2, Algorithm 3, and the baselines (we do not include the curve for FairBatch due to large overall loss and high variance in performance) under different values of $\gamma = [0.02,0.04,0.06,0.08,0.1]$ for five times and calculate the average BCE and the average BCE difference. We observe Algorithms 2 and 3 and the linear relaxation  have a similar  trade-off between $\hat{L}$ and $|\hat{L}_0-\hat{L}_1|$. 

\paragraph{Experiment with a non-linear model}
We repeat our first experiment with nonlinear models to demonstrate how we can use our algorithms to fine-tune a non-linear model. We work with the Law School Admission dataset, and we train a neural network with one hidden layer which consists of 125 neurons. We use sigmoid as the activation function for the hidden layer. 
We run the following algorithms,
\begin{itemize} [leftmargin=*,topsep=-1ex,itemsep=-0.2ex]
\item Penalty Method: We solve optimization problem \eqref{eq:penalty}. In this example, $\hat{L}$ and $\hat{L}_a$ are not convex anymore. The hyperparameters except for the learning rate remain the same as in the first experiment. The learning rate is set to be $0.001$. 
\item FairBatch: we train the whole network using FairBatch with mini-batch Adam optimization with a batch size of 100 and a learning rate of $0.001$.   
\item Linear Relaxation: In order to take advantage of CVXPY, first, we train the network without any fairness constraint using batch Adam optimization (i.e., the batch size is equal to the size of the training dataset) with a learning rate of $0.001$. Then, we fine-tune the parameters of the output layer. Note that the output layer has 126 parameters, and we fine-tune those under relaxed EL fairness. In particular, we solve  problem \eqref{eq:fineTune} after linear relaxation.
\item Algorithm 2 and Algorithm 3: We can run Algorithm 2 and Algorithm 3 to fine-tune the neural network. After training the network without any constraint using batch Adam optimization, we solve \eqref{eq:fineTune} using Algorithm 2 and Algorithm 3. 
\end{itemize}

Table \ref{Table3} illustrates the average and  standard deviation of empirical loss and the loss difference between Black and White students. 
Both Algorithm 2 and Algorithm 3 can achieve a fairness level (i.e., $|\hat{L}_0-\hat{L}_1|$) close to desired fairness level $\gamma$. Also, we can see that the MSE of Algorithm 2 and Algorithm 3 under the nonlinear model is slightly lower than the MSE under the linear model.  

\begin{figure}[t]
	\centering
		\includegraphics[width=0.75\linewidth]{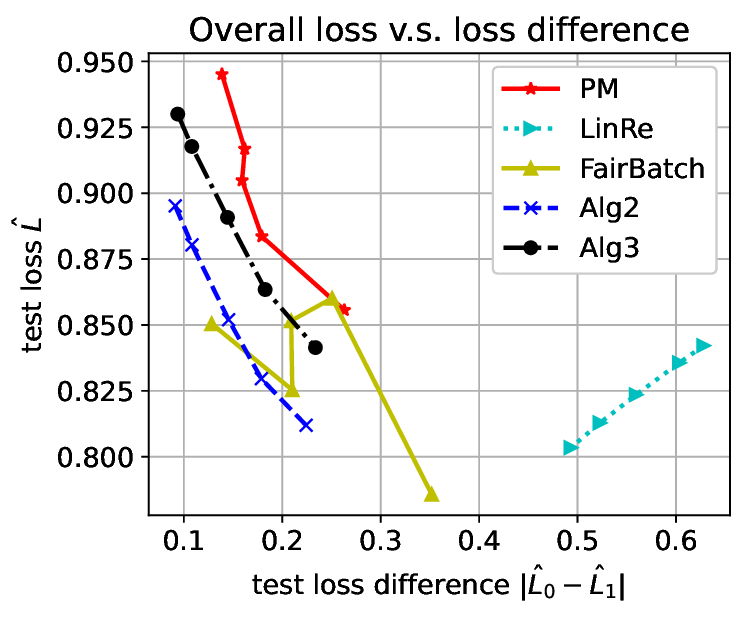}
		\caption{Trade-off between overall MSE and unfairness. A lower curve implies a better trade-off. }
		\label{fig:MSE_nonlinear}
	\vspace{-0.4cm}
\end{figure}

We also investigate how MSE $\hat{L}$ changes as a function of fairness level $|\hat{L}_1-\hat{L}_0|$. Figure \ref{fig:MSE_nonlinear} illustrates the MSE-fairness trade-off. To generate this plot, we repeat the experiment for $\gamma = [0.025,0.05,0.1,0.15,0.2]$. For each $\gamma$, we ran the experiment 5 times and calculated the average of MSE $\hat{L}$ and the average of MSE difference   using the test dataset. Based on Figure  \ref{fig:MSE_nonlinear}, we observe that FairBatch and LinRe are not very sensitive to the input $\gamma$. However, FairBatch may sometimes show a better trade-off than Algorithm 2. In this example, PM, Algorithm 2, and Algorithm 3 are very sensitive to $\gamma$, and as $\gamma$ increases, MSE $\hat{L}$ decreases and $|\hat{L}_0-\hat{L}_1|$ increases.

\begin{table}[t!]
\caption{Neural Network training  under {EL} fairness. The loss function in this example is the mean squared error loss.  } 
\centering
\resizebox{\linewidth}{!}{
\begin{tabular}{cccc} 
\toprule
& &$\gamma =0$ &  $\gamma = 0.1$ \\
\midrule
 &test loss 	&  $0.9490\pm 0.0584$ &  $0.9048\pm 0.0355$  \\
\rot{\rlap{~{PM}}} &test $|\hat{L}_0 - \hat{L}_1| $ 	&$0.1464\pm 0.1055$ & $0.1591\pm 0.0847$\\
\midrule
&  test loss  & $ 0.8489\pm 0.0195$ & $0.8235\pm  0.0165$ \\
\rot{\rlap{{\small{LinRe}}}} &test $|\hat{L}_0 - \hat{L}_1| $ & $0.6543\pm 0.0322$ &$0.5595\pm  0.0482$ \\
\midrule
&  test loss  & $0.9012\pm 0.1918$ & $0.8638\pm  0.0863$ \\
\rot{\rlap{~{\small{Fair}}}}
\rot{\rlap{~{\small{Batch}}}} &test $|\hat{L}_0 - \hat{L}_1| $ & $0.2771\pm 0.1252$ &$0.1491\pm  0.0928$ \\
\midrule
&test loss	& $0.9117\pm 0.0172$ &  $0.8519\pm 0.0195$  \\
\rot{\rlap{{ours}}}
\rot{\rlap{{Alg 2}}}&test $|\hat{L}_0-\hat{L}_1|$& $0.0761\pm 0.0498$ & $0.1454\pm 0.0749$ \\\midrule
 &test loss &$0.9427\pm 0.0190$&$0.8908\pm 0.0209$\\
 \rot{\rlap{{ours}}}
 \rot{\rlap{{Alg 3}}}&test $|\hat{L}_0-\hat{L}_1|$ &$0.0862\pm 0.0555$&$0.1423\pm0.0867$\\
\bottomrule 
\end{tabular}}
\label{Table3}

\end{table}

\textbf{Limitation and Negative Societal Impact.} 1) Our theoretical guarantees are valid under the stated assumptions (e.g., the convexity of $L({\w})$, i.i.d. samples, binary sensitive attribute). These assumptions have been clearly stated throughout this paper. 2) In this paper, we develop an algorithm for finding a fair predictor under EL fairness. However, we do not claim this notion is better than other fairness notions. Depending on the scenario, this notion may or may not be suitable for mitigating unfairness.  
\section{Conclusion}
In this work, we studied supervised learning problems under the Equalized Loss (EL) fairness \cite{zhang2019group}, a notion that requires the expected loss to be balanced  across different demographic groups. By imposing the EL constraint, the learning problem can be formulated as a non-convex  problem. 
 We proposed two algorithms with theoretical performance guarantees to find the global optimal and a sub-optimal solution to this non-convex problem. 
 \section*{Acknowledgment} This work is partially supported by the NSF under grants IIS-2202699, IIS-2301599, and ECCS-2301601. 
 
\bibliographystyle{plainnat}
\bibliography{main.bib}

\begin{thebibliography}{48}
\providecommand{\natexlab}[1]{#1}
\providecommand{\url}[1]{\texttt{#1}}
\expandafter\ifx\csname urlstyle\endcsname\relax
  \providecommand{\doi}[1]{doi: #1}\else
  \providecommand{\doi}{doi: \begingroup \urlstyle{rm}\Url}\fi

\bibitem[Abroshan et~al.()Abroshan, Khalili, and
  Elliott]{abroshan2022counterfactual}
Mahed Abroshan, Mohammad~Mahdi Khalili, and Andrew Elliott.
\newblock Counterfactual fairness in synthetic data generation.
\newblock In \emph{NeurIPS 2022 Workshop on Synthetic Data for Empowering ML
  Research}.

\bibitem[Abroshan et~al.(2023)Abroshan, Mishra, and
  Khalili]{abroshan2023symbolic}
Mahed Abroshan, Saumitra Mishra, and Mohammad~Mahdi Khalili.
\newblock Symbolic metamodels for interpreting black-boxes using primitive
  functions.
\newblock \emph{arXiv preprint arXiv:2302.04791}, 2023.

\bibitem[Agarwal et~al.(2018)Agarwal, Beygelzimer, Dud{\'\i}k, Langford, and
  Wallach]{agarwal2018reductions}
Alekh Agarwal, Alina Beygelzimer, Miroslav Dud{\'\i}k, John Langford, and Hanna
  Wallach.
\newblock A reductions approach to fair classification.
\newblock In \emph{International Conference on Machine Learning}, pages 60--69.
  PMLR, 2018.

\bibitem[Agarwal et~al.(2019)Agarwal, Dudik, and Wu]{agarwal2019fair}
Alekh Agarwal, Miroslav Dudik, and Zhiwei~Steven Wu.
\newblock Fair regression: Quantitative definitions and reduction-based
  algorithms.
\newblock In \emph{International Conference on Machine Learning}, pages
  120--129. PMLR, 2019.

\bibitem[Agrawal et~al.(2018)Agrawal, Verschueren, Diamond, and
  Boyd]{agrawal2018rewriting}
Akshay Agrawal, Robin Verschueren, Steven Diamond, and Stephen Boyd.
\newblock A rewriting system for convex optimization problems.
\newblock \emph{Journal of Control and Decision}, 5\penalty0 (1):\penalty0
  42--60, 2018.

\bibitem[Bartlett and Mendelson(2002)]{bartlett2002rademacher}
Peter~L Bartlett and Shahar Mendelson.
\newblock Rademacher and gaussian complexities: Risk bounds and structural
  results.
\newblock \emph{Journal of Machine Learning Research}, 3\penalty0
  (Nov):\penalty0 463--482, 2002.

\bibitem[Ben-Tal and Zibulevsky(1997)]{ben1997penalty}
Aharon Ben-Tal and Michael Zibulevsky.
\newblock Penalty/barrier multiplier methods for convex programming problems.
\newblock \emph{SIAM Journal on Optimization}, 7\penalty0 (2):\penalty0
  347--366, 1997.

\bibitem[Biega et~al.(2018)Biega, Gummadi, and Weikum]{biega2018equity}
Asia~J Biega, Krishna~P Gummadi, and Gerhard Weikum.
\newblock Equity of attention: Amortizing individual fairness in rankings.
\newblock In \emph{The 41st international acm sigir conference on research \&
  development in information retrieval}, pages 405--414, 2018.

\bibitem[Calmon et~al.(2017)Calmon, Wei, Vinzamuri, Ramamurthy, and
  Varshney]{calmon2017optimized}
Flavio~P Calmon, Dennis Wei, Bhanukiran Vinzamuri, Karthikeyan~Natesan
  Ramamurthy, and Kush~R Varshney.
\newblock Optimized pre-processing for discrimination prevention.
\newblock In \emph{Proceedings of the 31st International Conference on Neural
  Information Processing Systems}, pages 3995--4004, 2017.

\bibitem[Celis et~al.(2020)Celis, Keswani, and Vishnoi]{celis2020data}
L~Elisa Celis, Vijay Keswani, and Nisheeth Vishnoi.
\newblock Data preprocessing to mitigate bias: A maximum entropy based
  approach.
\newblock In \emph{International Conference on Machine Learning}, pages
  1349--1359. PMLR, 2020.

\bibitem[Conitzer et~al.(2019)Conitzer, Freeman, Shah, and
  Vaughan]{conitzer2019group}
Vincent Conitzer, Rupert Freeman, Nisarg Shah, and Jennifer~Wortman Vaughan.
\newblock Group fairness for the allocation of indivisible goods.
\newblock In \emph{Proceedings of the AAAI Conference on Artificial
  Intelligence}, volume~33, pages 1853--1860, 2019.

\bibitem[Dastin(2018)]{amazon}
Jeffrey Dastin.
\newblock Amazon scraps secret ai recruiting tool that showed bias against
  women.
\newblock \url{http://reut.rs/2MXzkly}, 2018.

\bibitem[Diamond and Boyd(2016)]{diamond2016cvxpy}
Steven Diamond and Stephen Boyd.
\newblock {CVXPY}: {A} {P}ython-embedded modeling language for convex
  optimization.
\newblock \emph{Journal of Machine Learning Research}, 17\penalty0
  (83):\penalty0 1--5, 2016.

\bibitem[Diana et~al.(2021)Diana, Gill, Kearns, Kenthapadi, and
  Roth]{diana2021minimax}
Emily Diana, Wesley Gill, Michael Kearns, Krishnaram Kenthapadi, and Aaron
  Roth.
\newblock Minimax group fairness: Algorithms and experiments.
\newblock In \emph{Proceedings of the 2021 AAAI/ACM Conference on AI, Ethics,
  and Society}, pages 66--76, 2021.

\bibitem[Donini et~al.(2018)Donini, Oneto, Ben-David, Shawe-Taylor, and
  Pontil]{donini2018empirical}
Michele Donini, Luca Oneto, Shai Ben-David, John~S Shawe-Taylor, and
  Massimiliano Pontil.
\newblock Empirical risk minimization under fairness constraints.
\newblock \emph{Advances in Neural Information Processing Systems}, 31, 2018.

\bibitem[Dressel and Farid(2018)]{COMPAS}
Julia Dressel and Hany Farid.
\newblock The accuracy, fairness, and limits of predicting recidivism.
\newblock \emph{Science advances}, 4\penalty0 (1):\penalty0 eaao5580, 2018.

\bibitem[Dwork et~al.(2012)Dwork, Hardt, Pitassi, Reingold, and
  Zemel]{dwork2012fairness}
Cynthia Dwork, Moritz Hardt, Toniann Pitassi, Omer Reingold, and Richard Zemel.
\newblock Fairness through awareness.
\newblock In \emph{Proceedings of the 3rd innovations in theoretical computer
  science conference}, pages 214--226, 2012.

\bibitem[Gupta and Kamble(2019)]{gupta2019individual}
Swati Gupta and Vijay Kamble.
\newblock Individual fairness in hindsight.
\newblock In \emph{Proceedings of the 2019 ACM Conference on Economics and
  Computation}, pages 805--806, 2019.

\bibitem[Hardt et~al.(2016)Hardt, Price, and Srebro]{hardt2016equality}
Moritz Hardt, Eric Price, and Nati Srebro.
\newblock Equality of opportunity in supervised learning.
\newblock \emph{Advances in neural information processing systems},
  29:\penalty0 3315--3323, 2016.

\bibitem[Harwell(2018)]{accent}
Drew Harwell.
\newblock The accent gap.
\newblock \url{http://wapo.st/3pUqZ0S}, 2018.

\bibitem[Jung et~al.(2019)Jung, Kearns, Neel, Roth, Stapleton, and
  Wu]{jung2019eliciting}
Christopher Jung, Michael Kearns, Seth Neel, Aaron Roth, Logan Stapleton, and
  Zhiwei~Steven Wu.
\newblock Eliciting and enforcing subjective individual fairness.
\newblock \emph{arXiv preprint arXiv:1905.10660}, 2019.

\bibitem[Kamiran and Calders(2012)]{kamiran2012data}
Faisal Kamiran and Toon Calders.
\newblock Data preprocessing techniques for classification without
  discrimination.
\newblock \emph{Knowledge and Information Systems}, 33\penalty0 (1):\penalty0
  1--33, 2012.

\bibitem[Khalili et~al.(2020)Khalili, Zhang, Abroshan, and
  Sojoudi]{khalili2020improving}
Mohammad~Mahdi Khalili, Xueru Zhang, Mahed Abroshan, and Somayeh Sojoudi.
\newblock Improving fairness and privacy in selection problems.
\newblock \emph{arXiv preprint arXiv:2012.03812}, 2020.

\bibitem[Khalili et~al.(2021)Khalili, Zhang, and Abroshan]{khalili2021fair}
Mohammad~Mahdi Khalili, Xueru Zhang, and Mahed Abroshan.
\newblock Fair sequential selection using supervised learning models.
\newblock \emph{Advances in Neural Information Processing Systems},
  34:\penalty0 28144--28155, 2021.

\bibitem[Kingma and Ba(2014)]{kingma2014adam}
Diederik~P Kingma and Jimmy Ba.
\newblock Adam: A method for stochastic optimization.
\newblock \emph{arXiv preprint arXiv:1412.6980}, 2014.

\bibitem[Kohavi(1996)]{kohavi1996scaling}
Ron Kohavi.
\newblock Scaling up the accuracy of naive-bayes classifiers: A decision-tree
  hybrid.
\newblock In \emph{Kdd}, volume~96, pages 202--207, 1996.

\bibitem[Komiyama et~al.(2018)Komiyama, Takeda, Honda, and
  Shimao]{komiyama2018nonconvex}
Junpei Komiyama, Akiko Takeda, Junya Honda, and Hajime Shimao.
\newblock Nonconvex optimization for regression with fairness constraints.
\newblock In \emph{International conference on machine learning}, pages
  2737--2746. PMLR, 2018.

\bibitem[Lohaus et~al.(2020)Lohaus, Perrot, and Von~Luxburg]{lohaus2020too}
Michael Lohaus, Michael Perrot, and Ulrike Von~Luxburg.
\newblock Too relaxed to be fair.
\newblock In \emph{International Conference on Machine Learning}, pages
  6360--6369. PMLR, 2020.

\bibitem[Lundberg and Lee(2017)]{lundberg2017shap}
Scott Lundberg and Su-In Lee.
\newblock A unified approach to interpreting model predictions.
\newblock \emph{arXiv preprint arXiv:1705.07874}, 2017.

\bibitem[Mahdavi et~al.(2012)Mahdavi, Yang, Jin, Zhu, and
  Yi]{mahdavi2012stochastic}
Mehrdad Mahdavi, Tianbao Yang, Rong Jin, Shenghuo Zhu, and Jinfeng Yi.
\newblock Stochastic gradient descent with only one projection.
\newblock \emph{Advances in neural information processing systems},
  25:\penalty0 494--502, 2012.

\bibitem[Nedi{\'c} and Ozdaglar(2009)]{nedic2009subgradient}
Angelia Nedi{\'c} and Asuman Ozdaglar.
\newblock Subgradient methods for saddle-point problems.
\newblock \emph{Journal of optimization theory and applications}, 142\penalty0
  (1):\penalty0 205--228, 2009.

\bibitem[Nemirovski(2004)]{nemirovski2004interior}
Arkadi Nemirovski.
\newblock Interior point polynomial time methods in convex programming.
\newblock \emph{Lecture notes}, 42\penalty0 (16):\penalty0 3215--3224, 2004.

\bibitem[Reimers et~al.(2021)Reimers, Bodesheim, Runge, and
  Denzler]{reimers2021towards}
Christian Reimers, Paul Bodesheim, Jakob Runge, and Joachim Denzler.
\newblock Towards learning an unbiased classifier from biased data via
  conditional adversarial debiasing.
\newblock \emph{arXiv preprint arXiv:2103.06179}, 2021.

\bibitem[Ribeiro et~al.(2016)Ribeiro, Singh, and Guestrin]{ribeiro2016lime}
Marco~Tulio Ribeiro, Sameer Singh, and Carlos Guestrin.
\newblock Why should i trust you?" explaining the predictions of any
  classifier.
\newblock In \emph{Proceedings of the 22nd ACM SIGKDD international conference
  on knowledge discovery and data mining}, pages 1135--1144, 2016.

\bibitem[Roh et~al.(2020)Roh, Lee, Whang, and Suh]{roh2020fairbatch}
Yuji Roh, Kangwook Lee, Steven~Euijong Whang, and Changho Suh.
\newblock Fairbatch: Batch selection for model fairness.
\newblock In \emph{International Conference on Learning Representations}, 2020.

\bibitem[Shalev-Shwartz and Ben-David(2014)]{shalev2014understanding}
Shai Shalev-Shwartz and Shai Ben-David.
\newblock \emph{Understanding machine learning: From theory to algorithms}.
\newblock Cambridge university press, 2014.

\bibitem[Shen et~al.(2022)Shen, Han, Cohn, Baldwin, and
  Frermann]{shen2022optimising}
Aili Shen, Xudong Han, Trevor Cohn, Timothy Baldwin, and Lea Frermann.
\newblock Optimising equal opportunity fairness in model training.
\newblock \emph{arXiv preprint arXiv:2205.02393}, 2022.

\bibitem[Wightman(1998)]{wightman1998lsac}
Linda~F Wightman.
\newblock Lsac national longitudinal bar passage study. lsac research report
  series.
\newblock 1998.

\bibitem[Williamson and Menon(2019)]{williamson2019fairness}
Robert Williamson and Aditya Menon.
\newblock Fairness risk measures.
\newblock In \emph{International Conference on Machine Learning}, pages
  6786--6797. PMLR, 2019.

\bibitem[Woodworth et~al.(2017)Woodworth, Gunasekar, Ohannessian, and
  Srebro]{woodworth2017learning}
Blake Woodworth, Suriya Gunasekar, Mesrob~I Ohannessian, and Nathan Srebro.
\newblock Learning non-discriminatory predictors.
\newblock In \emph{Conference on Learning Theory}, pages 1920--1953. PMLR,
  2017.

\bibitem[Wright(2001)]{wright2001convergence}
Stephen~J Wright.
\newblock On the convergence of the newton/log-barrier method.
\newblock \emph{Mathematical programming}, 90\penalty0 (1):\penalty0 71--100,
  2001.

\bibitem[Wu et~al.(2019)Wu, Zhang, and Wu]{wu2019convexity}
Yongkai Wu, Lu~Zhang, and Xintao Wu.
\newblock On convexity and bounds of fairness-aware classification.
\newblock In \emph{The World Wide Web Conference}, pages 3356--3362, 2019.

\bibitem[Zafar et~al.(2017)Zafar, Valera, Gomez~Rodriguez, and
  Gummadi]{zafar2017fairness}
Muhammad~Bilal Zafar, Isabel Valera, Manuel Gomez~Rodriguez, and Krishna~P
  Gummadi.
\newblock Fairness beyond disparate treatment \& disparate impact: Learning
  classification without disparate mistreatment.
\newblock In \emph{Proceedings of the 26th international conference on world
  wide web}, pages 1171--1180, 2017.

\bibitem[Zafar et~al.(2019)Zafar, Valera, Gomez-Rodriguez, and
  Gummadi]{zafar2019fairness}
Muhammad~Bilal Zafar, Isabel Valera, Manuel Gomez-Rodriguez, and Krishna~P
  Gummadi.
\newblock Fairness constraints: A flexible approach for fair classification.
\newblock \emph{The Journal of Machine Learning Research}, 20\penalty0
  (1):\penalty0 2737--2778, 2019.

\bibitem[Zhang et~al.(2018)Zhang, Lemoine, and Mitchell]{zhang2018mitigating}
Brian~Hu Zhang, Blake Lemoine, and Margaret Mitchell.
\newblock Mitigating unwanted biases with adversarial learning.
\newblock In \emph{Proceedings of the 2018 AAAI/ACM Conference on AI, Ethics,
  and Society}, pages 335--340, 2018.

\bibitem[Zhang et~al.(2019)Zhang, Khaliligarekani, Tekin, and
  Liu]{zhang2019group}
Xueru Zhang, Mohammadmahdi Khaliligarekani, Cem Tekin, and Mingyan Liu.
\newblock Group retention when using machine learning in sequential decision
  making: the interplay between user dynamics and fairness.
\newblock \emph{Advances in Neural Information Processing Systems},
  32:\penalty0 15269--15278, 2019.

\bibitem[Zhang et~al.(2020)Zhang, Khalili, and Liu]{zhang2020long}
Xueru Zhang, Mohammad~Mahdi Khalili, and Mingyan Liu.
\newblock Long-term impacts of fair machine learning.
\newblock \emph{Ergonomics in Design}, 28\penalty0 (3):\penalty0 7--11, 2020.

\bibitem[Zhang et~al.(2022)Zhang, Khalili, Jin, Naghizadeh, and
  Liu]{zhang2022fairness}
Xueru Zhang, Mohammad~Mahdi Khalili, Kun Jin, Parinaz Naghizadeh, and Mingyan
  Liu.
\newblock Fairness interventions as (dis) incentives for strategic
  manipulation.
\newblock In \emph{International Conference on Machine Learning}, pages
  26239--26264. PMLR, 2022.

\end{thebibliography}

\newpage
\appendix
\onecolumn

\section{Appendix}
\subsection{Some notes on the code for reproducibility}

In this part, we provide a description of the files provided in our GitHub repository.
\begin{itemize}
\item $law\_data.py$: This file includes a function called $law\_data(seed)$ which processes the law school admission dataset and splits the dataset randomly into training and test datasets (we keep 70\% of the datapoints for training). Later, in our experiments, we set the $seed$ equal to 0, 1, 2, 3, and 4 to get five different splits to repeat our experiments five times. 

\item $Adult\_data.py$: This file includes a function called $Adult\_dataset(seed)$ which processes the adult income dataset and splits the dataset randomly into training and test datasets. Later, in our experiments, we set the $seed$ equal to 0, 1, 2, 3, 4 to get five different splits to repeat our experiments five times. 

\item $Algorithms.py$: This file includes the following functions, 
\begin{itemize}
\item $ELminimizer(X0, Y0, X1, Y1, gamma, eta, model)$: This function implements \texttt{Elminimizer} algorithm. $(X0,Y0)$ are the training datapoints belonging to group $A=0$ and $(X1,Y1)$ are the datapoits belonging to group $A=1$. $gamma$ is the fairness level for EL constraint. $\eta$  is the reqularizer parameter (in our experiments, $\eta = 0.002$). $model$ determines the model that we want to train. If $model="linear"$, then we train a linear regression model. If $model="logistic"$, then we train a logisitic regression model. This function returns five variables $(w,b,l0,l1,l)$. $w,b$ are the weight vector and bias term of the trained model. $l_0,l_1$ are the average training loss of group 0 and group 1, respectively. $l$ is the overall training loss. 

\item $ Algorithm2(X0, Y0, X1, Y1, gamma, eta, model)$: This function implements Algorithm 2 which calls \texttt{Elminimizer} algorithm twice. This function also returns five variables $(w,b,l0,l1,l)$. These variables have been defined above. 

\item $ Algorithm3(X0, Y0, X1, Y1, gamma, eta, model)$: This function implements Algorithm 3 which finds a sub-optimal solution under EL fairness. This function also returns five variables $(w,b,l0,l1,l)$. These variables have been defined above.
\item $solve\_constrained\_opt(X0, Y0, X1, Y1, eta, landa, model)$: This function uses the CVXPY package to solve the optimization problem (4). We set $landa$ equal to $\lambda_{mid}^{(i)}$ to solve the optimization problem \eqref{eq:OptAlg} in iteration $i$ of Algorithm 1. 

\item $calculate\_loss(w,b,X0, Y0, X1, Y1, model)$: This function is used to find the test loss. $w,b$ are model parameters (trained by Algorithm 2 or 3). It returns the average loss of group 0 and group 1 and the overall loss based on the given dataset. 

\item $solve\_lin\_constrained\_opt(X0, Y0, X1, Y1, gamma, eta ,model)$: This function is for solving optimization problem \eqref{eq:approx} after linear relaxation. 
\end{itemize}

\item $Baseline.py$: this file includes the following functions,
\begin{itemize}
\item $penalty\_method(method, X\_0, y\_0, X\_1, y\_1, num\_itr ,lr,r,gamma,seed,epsilon)$ where $method$ can be either $"linear"$ for linear regression or $"logistic"$ for logistic regression. This function uses the penalty method and trains the model under EL using the Adam optimization. $num\_itr$ is the maximum number of iterations. $r$ is the regularization parameter (it is set to $0.002$ in our experiment). $lr$ is the learning rate and $gamma$ is the fairness level. $\epsilon$ is used for the stopping criterion. This function returns the trained model (which is a torch module), and training loss of group 0 and group 1, and the overall training loss.   
\item $fair\_batch(method, X\_0, y\_0, X\_1, y\_1, num\_itr ,lr,r,alpha,gamma,seed,epsilon)$: This function is used to simulate the FairBatch algorithm \cite{roh2020fairbatch}. The input parameters are similar to the input parameters of $penalty\_method$ except for $alpha$. This parameter determines how to adjust the sub-sampling distribution for mini-batch formation. Please look at the next section for more details. 
This function returns the trained model (which is a torch module), and training loss of group 0 and group 1, and the overall training loss.
\end{itemize}
\end{itemize}

$table1\_2.py$ uses the above functions to reproduce the results in Table 1 and Table 2. $figure1\_2.py$ uses the above functions to reproduce Figure 1 and Figure 2. We provide some comments in these files to make the code more readable. We have also provided code for training non-linear models. Please use $Table3.py$ and $figure3.py$ to generate the results in Table 3 and Figure 3, respectively.

Lastly, use the following command to generate results in Table 1:

\begin{itemize}
\item \texttt{python3 table1\_2.py --experiment=1 --gamma=0.0}
\item  \texttt{python3 table1\_2.py --experiment=1 --gamma=0.1}
\end{itemize}

Use the following command to generate results in Table 2:

\begin{itemize}
\item \texttt{python3 table1\_2.py --experiment=2 --gamma=0.0}
\item\texttt{python3 table1\_2.py --experiment=2 --gamma=0.1}
\end{itemize}

Use the following command to generate results in Table 3:

\begin{itemize}
\item \texttt{python3 table3.py  --gamma=0.0}
\item \texttt{python3 table3.py  --gamma=0.1}
\end{itemize}

Use the following command to generate results in Figure 1:

\begin{itemize}
\item \texttt{python3 figure1\_2.py --experiment=1}
\end{itemize}

Use the following command to generate results in Figure 2:

\begin{itemize}
\item \texttt{python3 figure1\_2.py --experiment=2}

\end{itemize}

Use the following command to generate results in Figure 3:
\begin{itemize}
\item \texttt{python3 figure3.py }

\end{itemize}

Note that you need to install packages in \texttt{requirements.txt}

\subsection{Notes on FairBatch \cite{roh2020fairbatch}} 
This method has been proposed to find a predictor under equal opportunity, equalized odd or statistical parity. In each epoch, this method identifies the disadvantaged group and increases the subsampling rate corresponding to the disadvantaged group in mini-batch selection for the next epoch. We modify this approach for $\gamma$-EL as follows, 
\begin{itemize}
\item We initialize the sub-sampling rate of group $a$ (denoted by $SR^{(0)}_a$) for mini-batch formation by $SR^{(0)}_a = \frac{n_a}{n}, a=0,1$. We Form the mini-batches using $SR^{(0)}_0$ and $SR^{(0)}_1$. 

\item At epoch $i$, we run gradient descent using the mini-batches formed by $SR^{(i-1)}_0$ and $SR^{(i-1)}_1$, and we obtain new model parameters $\pmb{w}_i$.
\item After epoch $i$, we calculate the empirical loss of each group. Then, we update $SR^{(i)}_a$ as follows, 
\begin{eqnarray*}
SR^{(i)}_a\longleftarrow SR^{(i-1)}_a+\alpha &&\mbox{if } \hat{L}_a(\w_i) - \hat{L}_{1-a}(\w_i)>\gamma \\
SR^{(i)}_a\longleftarrow SR^{(i-1)}_a-\alpha &&\mbox{if } \hat{L}_a(\w_i) - \hat{L}_{1-a}(\w_i)<-\gamma \\
SR^{(i)}_a\longleftarrow SR^{(i-1)}_a &&\mbox{o.w.},
\end{eqnarray*}
where is $\alpha$ is a hyperparameter and, in our experiment, is equal to $0.005$. 

\end{itemize}




\subsection{Details of numerical experiments and additional numerical results}
Due to the space limits of the main paper, we provide more details on our experiments here, 
\begin{itemize}
    \item Stopping criteria for penalty method and FairBatch: For stopping criteria, we stopped  the learning process when the change in the objective function is less than $10^{-6}$ between two consecutive  epochs. The reason that we used $10^{-6}$ was that we did not observe any significant change  by choosing a smaller value. 
    \item Learning rate for penalty method and FairBatch: We chose $0.005$ for the learning rate for training a linear model. For the experiment with a non-linear model, we set the learning rate to be $0.001$.   
    \item Stopping criteria for Algorithm 2 and Algorithm 3: As we stated in the main paper, we set $\epsilon = 0.01$ in \texttt{ELminimizer} and Algorithm 3. Choosing smaller $\epsilon$ did not change the performance significantly. 
    \item Linear Relaxation: Note that equation \eqref{eq:approx} after  linear relaxation is a convex optimization problem. We directly solve this optimization problem using CVXPY. 
\end{itemize}
The experiment has been done on a system with the following configurations: 24 GB of RAM, 2 cores of P100-16GB GPU, and 2 cores of Intel Xeon CPU@2.3 GHz processor. We used GPUs for training FairBatch. 

\subsection{Notes on the Reduction Approach \citep{agarwal2018reductions,agarwal2019fair}}
Let $Q(f)$ be a distribution over $\mathcal{F}$. In order to find optimal $Q(f)$ using the reduction approach, we have to solve the following optimization problem, 

\begin{eqnarray*}
\min_{Q} && \sum_{f\in\mathcal{F}} Q(f) \E\{l(Y,f(\pmb{X}))\} \\
s.t., && \sum_{f\in \mathcal{F}} Q(f) \E\{l(Y,f(\pmb{X}))|A=0\} = \sum_{f\in \mathcal{F}}Q(f) \E\{l(Y,f(\pmb{X}))\}\\
&& \sum_{f\in \mathcal{F}} Q(f) \E\{l(Y,f(\pmb{X}))|A=1\} = \sum_{f\in \mathcal{F}}Q(f) \E\{l(Y,f(\pmb{X}))\}
\end{eqnarray*}
Similar to \cite{agarwal2018reductions,agarwal2019fair}, we can re-write the above optimization problem in the following form, 
\begin{eqnarray*}
\min_{Q} && \sum_{f\in\mathcal{F}} Q(f) \E\{l(Y,f(\pmb{X}))\} \\
s.t., && \sum_{f\in \mathcal{F}} Q(f) \E\{l(Y,f(\pmb{X}))|A=0\} - \sum_{f\in \mathcal{F}}Q(f) \E\{l(Y,f(\pmb{X}))\}\leq 0 \\
&& -\sum_{f\in \mathcal{F}} Q(f) \E\{l(Y,f(\pmb{X}))|A=0\} + \sum_{f\in \mathcal{F}}Q(f) \E\{l(Y,f(\pmb{X}))\}\leq 0\\
&& \sum_{f\in \mathcal{F}} Q(f) \E\{l(Y,f(\pmb{X}))|A=1\} - \sum_{f\in \mathcal{F}}Q(f) \E\{l(Y,f(\pmb{X}))\}\leq 0 \\
&& -\sum_{f\in \mathcal{F}} Q(f) \E\{l(Y,f(\pmb{X}))|A=1\} + \sum_{f\in \mathcal{F}}Q(f) \E\{l(Y,f(\pmb{X}))\}\leq 0
\end{eqnarray*}
Then, the reduction approach forms the Lagrangian function as follows,
\begin{eqnarray*}
L(Q,\mu) &=& \sum_{f\in\mathcal{F}} Q(f) \E\{l(Y,f(\pmb{X}))\} \\ &-& \mu_1 \cdot \left( \sum_{f\in \mathcal{F}} Q(f) \E\{l(Y,f(\pmb{X}))|A=0\} - \sum_{f\in \mathcal{F}} Q(f) \E\{l(Y,f(\pmb{X}))\}\right)\\
&-& \mu_2 \cdot \left( -\sum_{f\in \mathcal{F}} Q(f) \E\{l(Y,f(\pmb{X}))|A=0\} + \sum_{f\in \mathcal{F}} Q(f) \E\{l(Y,f(\pmb{X}))\}\right)\\&-& \mu_3 \cdot  \left( \sum_{f\in \mathcal{F}} Q(f) \E\{l(Y,f(\pmb{X}))|A=1\} - \sum_{f\in \mathcal{F}} Q(f) \E\{l(Y,f(\pmb{X}))\}\right)\\
&-& \mu_4 \cdot \left( -\sum_{f\in \mathcal{F}} Q(f) \E\{l(Y,f(\pmb{X}))|A=1\} + \sum_{f\in \mathcal{F}} Q(f) \E\{l(Y,f(\pmb{X}))\}\right),\\
&&\mu_1\geq 0, \mu_2\geq 0, \mu_3\geq 0, \mu_4\geq 0.
\end{eqnarray*}

Since $f$ is parametrized with $\w$, we can find distribution $Q(\w)$ over $\mathbb{R}^{d_{\w}}$. Therefore, we rewrite the problem in the following form, 
\begin{eqnarray*}
L(Q(\w),\mu_1,\mu_2,\mu_3,\mu_4) &=& \sum_{\w} Q({\w}) L({\w}) \\ 
&-& \mu_1 \left( \sum_{{\w}} Q(\w) L_0({\w}) - \sum_{{\w}} Q(\w) L({\w})\right)\\
&-& \mu_2 \left( -\sum_{{\w}} Q(\w) L_0({\w}) + \sum_{{\w}} Q(\w) L({\w})\right)\\
&-& \mu_3 \left( \sum_{{\w}} Q(\w) L_1({\w}) - \sum_{{\w}} Q(\w) L({\w})\right)\\
&-& \mu_4 \left( -\sum_{{\w}} Q(\w) L_1({\w}) + \sum_{{\w}} Q(\w) L({\w})\right)
\end{eqnarray*}

The reduction approach  updates $Q({\w})$ and $(\mu_1,\mu_2,\mu_3,\mu_4)$  alternatively. Looking carefully at Algorithm 1 in \cite{agarwal2018reductions}, after updating $(\mu_1,\mu_2,\mu_3,\mu_4)$, we need to have access to an oracle that is able to solve the following optimization problem in each iteration, 
$$\min_{{\w}} (1+\mu_1-\mu_2+\mu_3-\mu_4)L({\w}) + (-\mu_1+\mu_2)L_0(\w)+(-\mu_3+\mu_4)L_1({\w}) $$

The above optimization problem is not convex for all $\mu_1,\mu_2,\mu_3,\mu_4$. Therefore, in order to use the reduction approach, we need to have access to an oracle that is able to solve the above non-convex optimization problem which is not available. Note that the original problem \eqref{eq:mainOP} is a non-convex optimization problem and using the reduction approach just leads to another non-convex optimization problem.

\subsection{Equalized Loss \& Bounded Group Loss}\label{sec:fairDef}
In this section, we study the relation between EL and BGL fairness notions. It is straightforward to see that any predictor satisfying $\gamma$-BGL also satisfies the   $\gamma$-EL. However, it is unclear to what extend an \textit{optimal} fair predictor under $\gamma$-EL  satisfies the BGL fairness notion. Next, we  theoretically  study the relation between BGL and EL fairness notions. 

Let $\pmb{w}^*$ be denoted as the solution to \eqref{eq:mainOP} and $f_{\pmb{w}^*}$ the corresponding optimal $\gamma$-{EL} fair predictor. Theorem \ref{Theo:2} below shows that under certain conditions, it is impossible for both groups to experience a loss larger than $2\gamma$ under the \textit{optimal} $\gamma$-{EL} fair predictor.

\begin{theorem}\label{Theo:2}
Suppose there exists a predictor that satisfies $\gamma$-BGL fairness notion. That is, the following optimization problem has at least one feasible point. 
\begin{eqnarray}\label{eq:BGLopt}
\min_{\pmb{w}} ~~L(\pmb{w}) \text{ s.t.}~~ L_a(\pmb{w})\leq\gamma, ~\forall a\in \{0,1\}.
\end{eqnarray}
Then, the followings hold,
\begin{eqnarray*}
\min\{L_0(\pmb{w}^*),L_1(\pmb{w}^*)\} &\leq& \gamma; \\
\max\{L_0(\pmb{w}^*),L_1(\pmb{w}^*)\} &\leq& 2\gamma.
\end{eqnarray*}
\end{theorem}
Theorem \ref{Theo:2} shows that $\gamma$-EL implies $2\gamma$-BGL if $\gamma$-BGL is a feasible constraint. Therefore, if $\gamma$ is not too small (e.g., $\gamma=0$), then EL and BGL are not contradicting each other. 

We emphasize that we are not claiming that whether {EL} fairness is better than {BGL}. Instead, these relations indicate the impacts the two fairness constraints could have on the model performance; the results may further provide the guidance for policy-makers.

\newpage

\subsection{Proofs}

In order to prove Theorem \ref{theo:main},  we first introduce two lemmas. 

\begin{lemma}\label{lem1}
Under assumption \ref{assump:2}, there exists $\overline{\pmb{w}} \in \mathbb{R}^{d_{\w}}$ such that $L_0(\overline{\pmb{w}}) = L_1(\overline{\pmb{w}}) = L(\overline{\pmb{w}})$ and $\lambda^{(0)}_{start} \leq L(\overline{\pmb{w}}) \leq \lambda^{(0)}_{end}$.
\end{lemma}

\textbf{Proof.} Let $q_0(\beta) = L_0((1-\beta)\pmb{w}_{G_0} + \beta \pmb{w}_{G_1})$ and  $q_1(\beta) = L_1((1-\beta)\pmb{w}_{G_0} + \beta \pmb{w}_{G_1})$, and $ q(\beta) = q_0(\beta) - q_1(\beta), \beta \in [0,1]$. Note that $\nabla_{\pmb{w}} L_a( \pmb{w}_{G_a}  ) = 0$ because $\pmb{w}_{G_a}$ is the minimizer of $L_a( \pmb{w}  )$. 

First, we show that $L_0((1-\beta)\pmb{w}_{G_0} + \beta \pmb{w}_{G_1})$ is an increasing function in $\beta$, and $L_1((1-\beta)\pmb{w}_{G_0} + \beta \pmb{w}_{G_1})$ is a decreasing function in $\beta$. Note that $q'_0(0) = (\pmb{w}_{G_1}-\pmb{w}_{G_0})^T  \nabla_{\pmb{w}} L_0( \pmb{w}_{G_0}  )  = 0$,  and $q_0(\beta)$ is convex because $L_0( \pmb{w})$ is convex. This implies that $q'(\beta)$ is an increasing function, and $q'_0(\beta) \geq 0, \forall \beta\in [0,1]$. Similarly, we can show that $q'_1(\beta)\leq 0, \forall \beta \in [0,1]$.  

Note that under Assumption \eqref{assump:2}, $q(0) <0$ and $q(1)>0$. Therefore, by the intermediate value theorem, the exists $\overline{\beta}\in (0,1)$ such that $q(\overline{\beta}) = 0$. Define $\overline{\pmb{w}} = (1-\overline{\beta})\pmb{w}_{G_0} + \overline{\beta}\pmb{w}_{G_1} $. We  have,

\begin{eqnarray*}
q(\overline{\beta}) &=& 0 \implies L_0(\overline{\pmb{w}}) = L_1(\overline{\pmb{w}}) = L(\overline{\pmb{w}})\\
\pmb{w}_{G_0} &\mbox{is}& \mbox{minimizer of } L_0 \implies\\ L(\overline{\pmb{w}})&=& L_0(\overline{\pmb{w}}) \geq \lambda^{(0)}_{start}\\
q'_0 (\beta) &\geq& 0, \forall \beta\in[0,1] \implies q_0(1) \geq q_0(\overline{\beta}) \implies\\ \lambda_{end}^{(0)} &\geq& L_0(\overline{\pmb{w}}) = L(\overline{\pmb{w}})
\end{eqnarray*}
 
 \begin{lemma}\label{lem2}
 $L_0(\pmb{w}_i^*) = \lambda_{mid}^{(i)}$, where $\pmb{w}_i^*$ is the solution to \eqref{eq:OptAlg}.
 \end{lemma}
 
 \textbf{Proof.} We proceed by contradiction. Assume that $L_0(\pmb{w}_i^*)<\lambda_{mid}^{(i)}$ (i.e., $\pmb{w}_i^*$ is an interior point of the feasible set of \eqref{eq:OptAlg}). Notice that $\pmb{w}_{G_1}$ cannot be  in the feasible set of \eqref{eq:OptAlg} because $L_0(\pmb{w}_{G_1}) = \lambda_{end}^{(0)} > \lambda_{mid}^{(i)}$. As a result, $\nabla_{\pmb{w}} L_1(\pmb{w}_i^*) \neq0$. This is a contradiction because $\pmb{w}_i^*$ is an interior point of the feasible set of a convex optimization and cannot be optimal if $\nabla_{\pmb{w}} L_1(\pmb{w}_i^*)$ is not equal to zero. 

\textbf{Proof [Theorem \ref{theo:main}]}

Now, we show that $L(\w^*)\in I_i$ for all $i$ ($\w^*$ is the solution to \eqref{eq:mainOP} when $\gamma=0$. As a result, $L_0(\w^*) = L_1(\w^*) = L(\w^*)$). Note that $L(\w^*)= L_0(\w^*) \geq \lambda_{start}^{(0)}$ because $\pmb{w}_{G_0}$ is the minimizer of $L_0$. Moreover, $\lambda_{end}^{(0)}\geq L(\w^*)$ otherwise $L(\overline{\pmb{w}})< L(\w^*)$ ($\overline{\pmb{w}}$ is defined in Lemma \ref{lem1}) and $\w^*$ is not optimal solution under 0-EL. Therefore, $L(\w^*)\in I_0$. 

Now we proceed by induction. Suppose $L(\w^*)\in I_i$. We show that $L(\w^*) \in I_{i+1}$ as well. We consider two cases. 
\begin{itemize}
\item $L(\w^*)\leq \lambda_{mid}^{(i)}$. In this case $\w^*$ is a feasible point for \eqref{eq:OptAlg}, and $L_1(\w_i^*)=\lambda^{(i)}\leq L_1(\w^*) = L(\w^*)\leq \lambda_{mid}^{(i)}$. Therefore, $L(\w^*)\in I_{i+1}$.

\item $L(\w^*)> \lambda_{mid}^{(i)}$. In this case, we proceed by contradiction to show that $\lambda^{(i)} \geq \lambda_{mid}^{(i)}$. Assume that $\lambda^{(i)}<\lambda_{mid}^{(i)}$. Define $r(\beta) = r_0(\beta) - r_1(\beta)$, where $r_a(\beta) = L_a((1-\beta)\pmb{w}_{G_0} + \beta \pmb{w}_i^*)$. Note that $\lambda^{(i)} = r_1(1)$ By Lemma \ref{lem2}, $r_0(1) = \lambda_{mid}^{(i)}$. Therefore, $r(1) = \lambda_{mid}^{(i)} - \lambda^{(i)}>0$. Moreover, under Assumption \ref{assump:2}, $r(0)<0$. Therefore, by the intermediate value theorem, there exists $\overline{\beta}_0\in (0,1)$ such that $r(\overline{\beta}_0) = 0$. Similar to the proof of Lemma \ref{lem1}, we can show that $r_0(\beta)$ in an increasing function for all $\beta \in [0,1]$. As a result $r_0(\overline{\beta}_0) <r_0 (1) = \lambda_{mid}^{(i)}$. Define $\overline{\pmb{w}}_0 = (1-\overline{\beta}_0)\pmb{w}_{G_0} + \overline{\beta}_0 \pmb{w}_i^*$. We have, 
\begin{eqnarray}
r_0(\overline{\beta}_0) = L_0(\overline{\pmb{w}}_0) = L_1(\overline{\pmb{w}}_0) = L(\overline{\pmb{w}}_0) < \lambda_{mid}^{(i)}\\
L(\w^*)> \lambda_{mid}^{(i)}
\end{eqnarray}
The last two equations imply that $\w^*$ is not a global optimal fair solution under $0$-EL fairness constraint and it is not the global optmal solution to \eqref{eq:mainOP}. This is a contradiction. Therefore, if $L(\w^*)> \lambda_{mid}^{(i)}$, then   $\lambda^{(i)} \geq \lambda_{mid}^{(i)}$. As a result, $L(\w^*) \in I_{i+1}$
\end{itemize} 

By two above cases and the nested interval theorem, we conclude that, 

$$ L(\w^*) \in \cap_{i=1}^{\infty} I_i, ~ \lim_{i\to \infty} \lambda_{mid}^{(i)} = L(\w^*), $$
$$ \mbox{define} \lambda_{mid}^{\infty} := \lim_{i\to \infty} \lambda_{mid}^{(i)}$$

Therefore, $\lim_{i\to \infty} \pmb{w}_i^*$ would be the solution to the following optimziation problem,

$$\arg\min_{\pmb{w}} L_1(\pmb{w}) s.t., L_0(\pmb{w}) \leq \lambda_{mid}^{\infty} = L(\w^*)$$

By lemma \ref{lem2}, the solution to above optimization problem (i.e., $\lim_{i\to \infty} \pmb{w}_i^*$) satisfies the following, $L_0(\lim_{i\to \infty} \pmb{w}_i^*) =  \lambda_{mid}^{\infty} = L(\w^*)$. Therefore, $\lim_{i\to \infty} \pmb{w}_i^*$ is the global optimal solution to optimization problem \eqref{eq:mainOP}.

\textbf{Proof [Theorem \ref{theo:gammaEL} ]}
Let's assume that $\w_O$ does not satisfy the $\gamma$-EL.\footnote{If $\w_O$ satisfies $\gamma$-EL, it will be the optimal predictor under $\gamma$-EL fairness. Therefore, there is no need to solve any constrained optimization problem.  Note that $\w_O$ is the solution to problem \eqref{eq:Unconstratint1}. } Let $\pmb{w}^*$ be the optimal weight vector under $\gamma$-EL. It is clear that $\pmb{w}^* \neq \w_O$.

\textbf{Step 1.} we show that one of the following holds, 
\begin{eqnarray}
L_0(\pmb{w}^*) - L_1(\pmb{w}^*) = \gamma \\
L_0(\pmb{w}^*) - L_1(\pmb{w}^*) = -\gamma 
\end{eqnarray}
Proof by contradiction. Assume $-\gamma < L_0(\pmb{w}^*) - L_1(\pmb{w}^*) <\gamma$. This implies that $\pmb{w}^*$ is an interior point of the feasible set of optimization problem \eqref{eq:mainOP}.  Since $\pmb{w}^* \neq \pmb{w}_{O}$, then $\nabla L(\pmb{w}^*) \neq 0$. As a result, object function of \eqref{eq:mainOP} can be improved at $\pmb{w}^*$ by moving toward $-\nabla L(\pmb{w}^*)$. This a contradiction. Therefore,  $|L_0(\pmb{w}^*) - L_1(\pmb{w}^*)| = \gamma$.

\textbf{Step 2.} Function $\pmb{w}_{\gamma} = \texttt{ELminimizer}(\pmb{w}_{G_0},\pmb{w}_{G_0},\epsilon,\gamma)$ is the solution to the following optimization problem, 
\begin{eqnarray}\label{eq:plusgamma}
 \min_{\pmb{w}} \Pr\{A=0\} L_0(\pmb{w}) + \Pr\{A=1\}L_1(\pmb{w}),\nonumber\\ s.t., L_0(\pmb{w}) - L_1(\pmb{w}) = \gamma
\end{eqnarray}
 
 To show the above claim, notice that the solution to optimization problem  \eqref{eq:plusgamma} is the same as the following, 
 
 \begin{eqnarray}\label{eq:plusgamma2}
 \min_{\pmb{w}} \Pr\{A=0\} L_0(\pmb{w}) + \Pr\{A=1\}\tilde{L}_1(\pmb{w}),\nonumber\\ s.t., L_0(\pmb{w}) - \tilde{L}_1(\pmb{w}) = 0,
\end{eqnarray}
where $\tilde{L}_1(\w) = L_1(\w) + \gamma$.
Since $L_0(\w_{G_0}) - \tilde{L}_1(\w_{G_0})<0 $ and $L_0(\w_{G_1}) - \tilde{L}_1(\w_{G_1})>0 $, by Theorem \ref{theo:main}, we know that $\w_{\gamma} = \texttt{ELminimizer}(\pmb{w}_{G_0},\pmb{w}_{G_0},\epsilon,\gamma)$ find the solution to \eqref{eq:plusgamma2} when $\epsilon$ goes to zero. 

Lastly, because $|L_0(\pmb{w}^*) - L_1(\pmb{w}^*)| = \gamma$, we have, 

\begin{eqnarray}
\w^* = \left\lbrace \begin{array}{ll} \w_\gamma & \mbox{if } L(\w_\gamma) \leq L(\w_{-\gamma})\\
\w_{-\gamma} & \mbox{o.w.}\end{array} \right.
\end{eqnarray}
Thus, Algorithm \ref{Alg2} finds the solution to \eqref{eq:mainOP}. 

\textbf{Proof [Theorem \ref{Theo:3}]}

\begin{enumerate}
\item Under Assumption \ref{assump:2}, $g(1)<0$. Moreover, $g(0) \geq 0$. Therefore, by the intermediate value theorem, there exists $\beta_0\in [0,1]$ such that $g(\beta_0) = 0$. 

\item Since $\w_O$ is the minimizer of $L(\w)$, $h'(0) = 0$. Moreover, since $L(\w)$ is strictly convex, $h(\beta)$ is strictly convex and $h'(\beta)$ is strictly increasing function. As a result, $h'(\beta)>0$ for  $\beta>0$, and $h(\beta)$ is strictly increasing.

\item Similar to the above argument, $ s(\beta) = L_{\hat{a}}((1-\beta) \pmb{w}_O + \beta \w_{G_{\hat{a}}}) $ is strictly decreasing function (notice that $s'(1) =0$ and $s(\beta)$ is strictly convex). 

Note that since $h(\beta) = \Pr\{A=\hat{a}\} L_{\hat{a}}((1-\beta) \pmb{w}_O + \beta \w_{G_{\hat{a}}}) + \Pr\{A=1-\hat{a}\} L_{1-\hat{a}}((1-\beta) \pmb{w}_O + \beta \w_{G_{\hat{a}}})$ is strictly increasing and $ L_{\hat{a}}((1-\beta) \pmb{w}_O + \beta \w_{G_{\hat{a}}}) $ is strictly decreasing. Therefore, we conclude that $L_{1-\hat{a}}((1-\beta) \pmb{w}_O + \beta \w_{G_{\hat{a}}})$ is strictly increasing. As a result, $g(\beta)$ should be strictly decreasing.

\end{enumerate}

\textbf{Proof [Theorem \ref{theo:sub}]}
First, we show that if $g_{\gamma}(0) \leq 0$, then $\w_O$ satisfies $\gamma$-{EL}.

$$g_{\gamma}(0)\leq 0 \implies g(\beta) - \gamma \leq 0 \implies L_{\hat{a}}(\w_O) - L_{1-\hat{a}}(\w_O) \leq \gamma$$

Moreover, $L_{\hat{a}}(\w_O) - L_{1-\hat{a}}(\w_O)\geq 0$ because $\hat{a} = \arg\max_a L_a(\w_O)$.  Therefore, $\gamma$-{EL} is satisfied. 

Now, let $g_{\gamma}(0) > 0$. Note that under Assumption \ref{assump:2}, $g_{\gamma}(1) = L_{\hat{a}}(\w_{G_{\hat{a}}}) - L_{1-\hat{a}}(\w_{G_{\hat{a}}}) -\gamma <0 $. Therefore, by the intermediate value theorem, there exists $\beta_0$ such that $g_{\gamma}(\beta_0) = 0$. Moreover, based on Theorem \ref{theo:sub}, $g_{\gamma}$ is a strictly decreasing function. Therefore, the binary search proposed in Algorithm 3 converges to the root of $g_{\gamma}(\beta)$.  As a result, $(1-\beta_{mid}^{(\infty)})\w_O + \beta_{mid}^{(\infty)} \w_{G_{\hat{a}}}$ satisfies  $\gamma$-{EL}. Note that since $g(\beta)$ is strictly decreasing, and $g(\beta_{mid}^{(\infty)}) = \gamma$, and $\beta^{(\infty)}_{mid}$ is the smallest possible $\beta$ under which $(1-\beta)\w_O + \beta \w_{G_{\hat{a}}}$ satisfies $\gamma$-{EL}. Since $h$ is increasing,  the smallest possible  $\beta$ gives us a better accuracy.

\textbf{Proof [Theorem \ref{theo:sub_perf}]}
If $g_{\gamma}(0) \leq 0$, then $\w_O$ satisfies $\gamma$-{EL}, and $\underline{\w} = \w_O$. In this case, it is easy to see that $L(\w_O) \leq \max_{a\in \{0,1\}} L_a(\w_O)$ (because  $L(\w_O)$ is a weighted average of $L_0(\w_O)$ and $L_1(\w_O))$. 

Now assume that $g_{\gamma}(0) >0$. Note that if we prove this theorem for $\gamma=0$, then the theorem will hold for $\gamma>0$. This is because the \textbf{optimal} predictor under $0$-EL satisfies $\gamma$-EL condition as well. In other words, $0$-EL is a stronger constraint than $\gamma$-EL.

Let $\gamma = 0$. In this case, Algorithm 3 finds $\underline{\w} = (1-\beta_0)\w_O + \beta_0 \w_{G_{\hat{a}}}$, where $\beta_0$ is defined in Theorem \ref{Theo:3}. We have,

$$(*)~~ g(\beta_0) = 0 = L_{\hat{a}}(\underline{\w}) - L_{1-\hat{a}}(\underline{\w})   $$
In the proof of theorem \ref{Theo:3}, we showed that $L_{\hat{a}}((1-\beta) \pmb{w}_O + \beta \w_{G_{\hat{a}}})$ is decreasing in $\beta$. Therefore, we have, 
$$ (**)~~ L_{\hat{a}}(\underline{\w}) \leq L_{\hat{a}}(\w_O)  $$

Therefore, we have,
\begin{eqnarray}
L(\underline{\w}) &=& \Pr(A=0)\cdot L_{\hat{a}}(\underline{\w})  + (1-\Pr(A=1))\cdot L_{1-\hat{a}}(\underline{\w})  \\ (\mbox{By }(*))~~~~~~~~~~~~~~~~ &=& L_{\hat{a}}(\underline{\w}) \\(\mbox{By } (**))~~~~~~~~~~~~~~~ &\leq &L_{\hat{a}}(\w_O)
\end{eqnarray}

\textbf{Proof [Theorem \ref{theo:learnable}]} 

By the triangle inequality, the following holds, 

\begin{eqnarray}
\sup_{f_{\pmb{w}}\in \mathcal{F}} | |L_0(\pmb{w}) - L_1(\pmb{w}) | - |\hat{L}_0(\pmb{w}) - \hat{L}_1(\pmb{w})| | \leq \\ \sup_{f_{\pmb{w}}\in \mathcal{F}} |L_0(\pmb{w}) - \hat{L}_0(\pmb{w})| + \sup_{f_{\pmb{w}}\in \mathcal{F}} |L_1(\pmb{w}) - \hat{L}_1(\pmb{w})|.
\end{eqnarray}
Therefore, with probability at least $1-2\delta$ we have, 
 \begin{eqnarray}
 \label{eq:ineq}
 \sup_{f_{\pmb{w}}\in \mathcal{F}} | |L_0(\pmb{w}) - L_1(\pmb{w}) | - |\hat{L}_0(\pmb{w}) - \hat{L}_1(\pmb{w})| | \leq \nonumber \\ B(\delta,n_0,\mathcal{F}) + B(\delta,n_1,\mathcal{F})
 \end{eqnarray}
 As a result, with probability $1-2\delta$ holds,
 \begin{eqnarray}\label{eq:subset}
 \{\pmb{w}| f_{\pmb{w}} \in \mathcal{F}, |L_0(\pmb{w}) - L_1(\pmb{w}) |\leq \gamma\} \subseteq\nonumber \\  \{\pmb{w}| f_{\pmb{w}} \in \mathcal{F}, |\hat{L}_0(\pmb{w}) - \hat{L}_1(\pmb{w}) |\leq \hat{\gamma}\}
 \end{eqnarray}
 Now consider the following, 

 \begin{equation}
L(\pmb{\hat{w}}) -  L(\pmb{w}^*) = L(\pmb{\hat{w}}) - \hat{L}(\pmb{\hat{w}}) + \hat{L}(\pmb{\hat{w}}) - \hat{L}(\pmb{w}^*) +  \hat{L}(\pmb{w}^*) - L(\pmb{w}^*)
 \end{equation}
 By \eqref{eq:subset}, $\hat{L}(\pmb{\hat{w}}) - \hat{L}(\pmb{w}^*) \leq 0$ with probability $1-2\delta$. Thus, with probability at least $1-2\delta$, we have, 
 
\begin{equation}
    L(\pmb{\hat{w}}) -  L(\pmb{w}^*) \leq  L(\pmb{\hat{w}}) - \hat{L}(\pmb{\hat{w}})  +  \hat{L}(\pmb{w}^*) - L(\pmb{w}^*).
\end{equation}
Therefore, under assumption \ref{assump:learnable}, we can conclude with probability at least $1-6\delta$, $L(\pmb{\hat{w}}) -  L(\pmb{w}^*) \leq 2B(\delta,n,\mathcal{F})$. In addition, by \eqref{eq:ineq}, with probability at least $1-2\delta$, we have, 

\begin{eqnarray*}
 |L_0(\pmb{\hat{w}}) - L_1(\pmb{\hat{w}})| &\leq& B(\delta,n_0,\mathcal{F}) + B(\delta,n_1,\mathcal{F}) + |\hat{L}_0(\pmb{w}) - \hat{L}_1(\pmb{w})|  \nonumber \\
 &\leq& \hat{\gamma} + B(\delta,n_0,\mathcal{F}) + B(\delta,n_1,\mathcal{F})\\ &=& \gamma + 2B(\delta,n_0,\mathcal{F}) + 2B(\delta,n_1,\mathcal{F})
\end{eqnarray*}

\textbf{Proof [Theorem \ref{Theo:2}]}
Let $\pmb{\tilde{w}}$ be a feasible point of optimization problem \eqref{eq:BGLopt}, then  $\pmb{\tilde{w}}$ is also a feasible point to \eqref{eq:mainOP}. 

We proceed by contradiction. We consider three cases,

\begin{itemize}
\item If $\min\{L_0(\pmb{w}^*),L_1(\pmb{w}^*)\} > \gamma$ and $\max\{L_0(\pmb{w}^*),L_1(\pmb{w}^*)\} > 2\gamma$. In this case,  $$L(\pmb{w}^*) > \gamma \geq L(\pmb{\tilde{w}}). $$ This is a contradiction because it implies that $\pmb{w}^*$ is not an optimal solution to \eqref{eq:mainOP}, and $\tilde{{\w}}$ is a better solution for \eqref{eq:mainOP}.
\item If $\min\{L_0(\pmb{w}^*),L_1(\pmb{w}^*)\} > \gamma$ and $\max\{L_0(\pmb{w}^*),L_1(\pmb{w}^*)\} \leq 2 \gamma$. This case is similar to above. $\min\{L_0(\pmb{w}^*),L_1(\pmb{w}^*)\} > \gamma$ implies that $L(\pmb{w}^*) > \gamma \geq L(\pmb{\tilde{w}}). $ This is a contradiction because it implies that $\pmb{w}^*$ is not an optimal solution to \eqref{eq:mainOP}. 
\item If $\min\{L_0(\pmb{w}^*),L_1(\pmb{w}^*)\} \leq \gamma$ and $\max\{L_0(\pmb{w}^*),L_1(\pmb{w}^*)\} > 2\cdot \gamma$.
We have:  $$\max\{L_0(\pmb{w}^*),L_1(\pmb{w}^*)\} - \min\{L_0(\pmb{w}^*),L_1(\pmb{w}^*)\}  > \gamma,$$ which shows that $\pmb{w}^*$ is not a feasible point for \eqref{eq:mainOP}. This is a contradiction.

\end{itemize}

Therefore,  $\max\{L_0(\pmb{w}^*),L_1(\pmb{w}^*)\} \leq 2\gamma$ and $\min\{L_0(\pmb{w}^*),L_1(\pmb{w}^*)\} \leq \gamma$.

\end{document}